# Effective Predictive Modeling for Emergency Department Visits and Evaluating Exogenous Variables Impact: Using Explainable Meta-learning Gradient Boosting

Mehdi Neshat[1,2], Michael Phipps[1], Nikhil Jha[3,4], Danial Khojasteh[5], Michael Tong[6], Amir Gandomi[2,7]

[1] *Canberra Health Services, Canberra Hospital, Canberra, ACT 2605, Australia mehdi.neshat@ieee.org; michael.phipps@act.gov.au*

[2] *Faculty of Engineering and Information Technology, University of Technology Sydney, Ultimo, Sydney, 2007, NSW, Australia neshat.mehdi@uts.edu.au, amir.gandomi@uts.edu.au*

[3]*Research Operations and Clinical Trials, Canberra Health Service, Canberra Hospital, Canberra, 2605, ACT, Australia nikhil.jha@act.gov.au*

[5]*School of Civil and Environmental Engineering, University of New South Wales, Sydney, Australia danial.khojasteh@unsw.edu.au*

[6]*National Centre for Epidemiology and Population Health, ANU College of Health and Medicine, The Australian National University, Canberra, ACT 2601, Australia*
*Michael.Tong@anu.edu.au*

[7]*University Research and Innovation Center (EKIK), Obuda University, Budapest, 1034, Hungary*

---

**Abstract**

**Objective**: Over an extensive duration, administrators and clinicians have endeavoured to predict Emergency Department (ED) visits with precision, aiming to optimise resource distribution through adept adjustments like upstaffing or down-staffing, modifying Operating Room (OR) timetables, and foretelling the demand for substantial resources. Despite the proliferation of diverse AI-driven models tailored for precise prognostication, this task persists as a formidable challenge, besieged by constraints such as restrained generalisability, susceptibility to overfitting and underfitting, scalability issues, and complex fine-tuning hyper-parameters. Furthermore, integrating climate parameters into predictive models could potentially enhance the accuracy and robustness of ED visit predictions, providing valuable insights for effective resource management in healthcare settings.
**Methods**: In this study, aiming to overcome the constraints of prevailing AI-driven frameworks, we introduce a novel Meta-learning Gradient Booster (Meta-ED) approach for precisely forecasting daily ED visits. Leveraging a comprehensive dataset of exogenous variables, including socio-demographic characteristics, healthcare service use, chronic diseases, diagnosis, and climate parameters spanning 23 years from Canberra Hospital in ACT, Australia. The proposed Meta-ED consists of four foundational learners—Catboost, Random Forest, Extra Tree, and lightGBoost—alongside a dependable top-level learner, Multi-Layer Perceptron (MLP), by combining the unique capabilities of varied base models (sub-learners), Meta-ED endeavours to strengthen predictive precision and robustness concerning ED attendance patterns.
**Results**: Our study assesses the efficacy of the Meta-ED model through an extensive comparative analysis involving 23 distinct models, encompassing a diverse range of architectures such as sequential models, neural networks, treebased algorithms, and ensemble methods. The evaluation outcomes reveal a notable superiority of Meta-ED over the other models in terms of accuracy at 85.7% (95% CI [85.4%, 86.0%]) and across a spectrum of 10 evaluation metrics. Notably, when compared with prominent techniques, including XGBoost, Random Forest (RF), AdaBoost, Light-GBoost, and Extra Tree (ExT), Meta-ED showcases substantial accuracy enhancements of 58.6%, 106.3%, 22.3%, 7.0%, and 15.7%, respectively. Furthermore, incorporating weather-related features demonstrates a 3.25% improvement in the prediction accuracy of visitors' numbers, which is pivotal in capturing seasonal trends that influence fluctuations in patient volumes. This underscores the model's adeptness in discerning nuanced patterns within the dataset.
**Conclusions**: The encouraging outcomes of our study underscore Meta-ED as a foundation model for the precise prediction of daily ED visitors. This highlights its potential as a reliable and robust framework in predictive analytics, offering a promising trajectory for enhancing prediction accuracy in complex and dynamic healthcare environments, particularly in the realm of predicting sequential data characterised by heterogeneous features and constrained datasets.

---

## 1. Introduction

The Emergency Department (ED), distinct from other hospital units, faces challenges in efficiently managing resources due to its inability to regulate patient influx [1]. Serving as the hospital's primary point of entry, the ED confronts time constraints in responding to fluctuations in patient volume [2]. Swift and critical care within the emergency medical systems is expanding. Yet, the persistent issue of ED overcrowding poses a notable risk to patient welfare, particularly for those with chronic ailments and infectious conditions [3, 4]. ED congestion leads to delays in patient transfer and financial setbacks for hospitals as they divert cases to alternate facilities [5]. Precisely forecasting ED visitation rates is pivotal for optimising resource distribution and enabling proactive staffing adjustments, operational modifications, and resource planning [6].

Forecasting the number of ED visitors presents a tough challenge primarily because no singular factor exhibits the precision essential for day-to-day or weekly planning [7, 8]. Therefore, a holistic multivariate approach becomes imperative to identify the pivotal predictive elements, encompassing demographic variables, historical ED visit trends, meteorological data [9], air quality metrics [10], major events, and more. Currently, most healthcare institutions rely on basic algorithms for short-term forecasting of emergency admissions, utilising elementary rolling averages for daily estimations [11]. An uncomplicated approach involves enhancements through the integration of Bayesian methodologies or autoregressive inductive moving averages, incorporating diverse datasets [12, 13]. Nonetheless, Bayesian-driven methodologies may pose computational challenges, particularly with intricate models or extensive data sets, potentially constraining their real-time application feasibility.

Machine learning (ML) models emerge as a more attractive method for predicting ED presentations [14] than Bayesian and classical prediction approaches (like linear regression [15], Naive Bayes [16]), Support Vector Regression (SVR) [17, 18], and Nonlinear Least Square Regression (NLSR) [19] thanks to their adaptability in managing intricate relationships and various data types. This adaptability allows a superior grasp of the complex patterns and interdependencies inherent in ED visitor data.

In the field of ML models, deep learning (DL) methodologies have risen to prominence due to their exceptional ability to unravel complex interconnections between variables and results. This makes them particularly effective at forecasting ED attendance. Unlike traditional models, deep learning strategies can intuitively detect intricate temporal patterns and relationships that define patient arrivals, providing more precise and flexible solutions. Some researchers [20, 17] have highlighted the success of deep learning frameworks, including Long Short-Term Memory (LSTM), Convolutional Neural Networks (CNNs), and stacked structures that integrate Gated Recurrent Units (GRUs) and Recurrent Neural Networks (RNNs). These sophisticated architectures excel at modelling patient flow dynamics, particularly in cases involving nonlinear and high-dimensional datasets, underscoring their effectiveness in handling complex data.

As demonstrated in these investigations, the adaptability of deep learning models is a key factor in their superiority over traditional ML techniques such as Random Forest (RF) [21, 22] and XGBoost in forecasting ED demand. For instance, studies [17] found that a deep-stacked framework and pre-trained language models [23] are able to effectively capture hidden temporal characteristics that other models might miss. Importantly, the model's accuracy seemed to be less affected by hyper-parameters like the dimensions of hidden layers or the count of hidden units, suggesting that the strength of these models lies more in their architecture than in the meticulous adjustment of specific parameters. This adaptability makes deep learning particularly suitable for the complex, unpredictable landscape of ED patient flow. However, it is crucial to recognise the difficulties in making direct comparisons across various studies due to discrepancies in data sources, regional patient profiles, and data accessibility. Despite these hurdles, the prevailing evidence indicates that deep learning models represent a formidable asset for advancing ED management, empowering healthcare institutions to allocate resources more adeptly and enhance patient care. Attention mechanisms [24] have remarkably propelled the field of time series forecasting forward, particularly with the advent of the Transformer model. In contrast to conventional sequential deep learning architectures, attention facilitates the elimination of recurrence and memory components by evaluating the entire input sequence simultaneously, pinpointing crucial elements for enhanced predictions. The self-attention feature in Transformers empowers them to adeptly grasp dependencies among distant elements in a sequence, rendering them exceptionally well-suited for managing extensive datasets. Recently, a few studies investigated the application of attention [25] and Transformers [26] to tackle the intricacies of forecasting ED attendance, highlighting their promise in unravelling temporal patterns and interconnections within health-related data.

The integration of external factors, often referred to as exogenous variables, to significantly enhance the accuracy of predicting the number of visitors to ED has emerged as a prevalent and effective strategy in a multitude of recent research endeavours. This innovative approach, where a variety of influences, including but not limited to climatic conditions and calendar-related information, have been extensively employed, as evidenced by the notable contributions made by researchers [25, 22, 14]. For example, regarding the application of exogenous variables, a recent study [25] developed a deep learning model with an Attention layer to predict ED admissions and achieved promising results at 80% of R2. Alvarez et al. [25] compared with a DNN and a Naive baseline, demonstrating superior performance across all cases. Among the exogenous inputs, calendar data proved the most valuable, followed by weather and air quality, while Google Trends data reduced model accuracy. The Attention model is also smaller and more computationally efficient than the LSTM-based model. Another study [20] meticulously incorporated various weather-related parameters, which encompassed elements like atmospheric temperature, wind velocity, the directional flow of winds, and the extent of cloud visibility while also integrating calendar-based data that featured significant time markers such as public holidays and academic calendars. In addition to this, other scholarly works [27, 28, 29] similarly adopted a range of external variables to inform their analyses, whereas certain studies. Beyond these factors, a broader spectrum of variables, such as air quality metrics (as examined by [30]), socioeconomic determinants (as highlighted in the research by [31]), and the prevalence of influenza outbreak levels (discussed in the study by [32]), have also been employed to refine the predictive models. Collectively, these diverse and innovative studies underscore the significant value and importance of incorporating a wide array of exogenous data points in order to substantially enhance the overall accuracy of forecasting visitor numbers in emergency departments. However, combining a wide array of parameters results in heterogeneous data that poses significant challenges for conventional predictive models, along with the current ML and Deep Learning (DL) frameworks [33], in accurately forecasting short-term or long-term emergency department traffic due to these variables' intricate nature and interactions.

Based on our literature review, hybrid ML models [34, 6, 35] have demonstrated superior performance over standalone models in predicting ED visitor numbers by combining the strengths of several algorithms. For instance, linear regression models are effective at capturing overall trends, whereas non-linear models, such as NNs or DLs, are adept at identifying complex interactions and relationships. Integrating these methodologies produces a model capable of both generalising effectively and capturing intricate patterns in the data. One recent study [36] introduced a hybrid ML method for predicting ED attendance by combining linear regression, neural networks (NNs), and regression tree ensembles (RTE). The approach uses external variables like socio-economic, weather, and temporal data to improve weekly forecasts. Trained on 11 years of data and tested over one year sequentially, the model achieved around 5% in validation error, showing its potential for enhancing ED management. Furthermore, Lopez et al. [36] reported that external data such as climate and socio-economic information could not significantly enhance the accuracy of the prediction.

This study presents a pioneering solution to the challenge of forecasting daily Emergency Department (ED) visitor numbers. The Meta-ED model, a novel meta-ensemble model, is introduced as an advanced hybrid machine learning technique. It is designed to optimize the aggregation of predictions from multiple sub-models and is trained on real, heterogeneous data from Canberra Hospital, ACT, Australia. The use of real, heterogeneous data from a healthcare setting enhances the model's credibility and applicability to real-world scenarios. The key contributions of this work are:

- This study establishes a comprehensive comparative framework for predicting ED attendance. It incorporates 23 widely recognised machine learning, sequential deep learning, and ensemble methods, ensuring a meticulous and thorough evaluation of the Meta-ED model's capabilities.
- A new, highly effective predictive model is introduced, which combines an optimal selection of sub-learners. These sub-learners, chosen based on their superior comparative performance, include two boosting methods (CatBoost and LightGBM) and two tree-based methods (Random Forest and Extra Trees), with a Master-learner (MLP) used to enhance prediction accuracy.
- The architecture and hyper-parameters of the proposed meta-ensemble model are optimized using an efficient differential evolution method paired with a fast local search algorithm (Nelder-Mead).
- Technical explainability experiments (XAI) are conducted to assess and compare the importance of exogenous features in ED visitor prediction, utilising both local and global interpretation techniques.
- The performance of the Meta-learner model is rigorously evaluated using several performance metrics. These metrics provide a comprehensive understanding of the model's performance and its advantages over existing models. It is demonstrated that the proposed Meta-learner outperforms the other 23 predictive models in both accuracy and robustness.

In the next section, we present an overview of the methodology, including introducing various machine learning and sequential deep learning techniques, alongside the technical specifics of our proposed meta-learning approach. This includes key aspects such as feature selection, hyper-parameter optimisation, explainable AI and our dataset characteristics and analysis, as discussed in **Section 2**. Following this, we provide a detailed presentation of the numerical results derived from the aforementioned methods and engage in a thorough analysis of the findings in **Section 3**. This allows for a clear comparison of the performance and efficiency of the proposed framework. Lastly, we conclude the paper by highlighting the key outcomes and emphasising the benefits of our method in comparison to existing solutions, as discussed in **Section 4**.

## 2. Methods and Materials

In this section, we introduce the collected datasets used in the prediction modelling, such as ED admissions, diagnosis, demographic records, and climate and temporal features. Next, we will introduce and describe the technical details of MLs, DLs, and Ensemble methods developed and compared in this study, as well as our proposed solution to improve the prediction accuracy of ED visitors.

### *2.1. Emergency Department Data*

The Emergency Department at Canberra Hospital plays a vital role in delivering healthcare services to the population of the Australian Capital Territory (ACT). This study leverages the DHR dataset, sourced from the ED, to explore patient demographics, treatment patterns, and outcomes. The dataset spans 23 years, from January 1999 to December 2022, offering a comprehensive foundation for detailed analysis. Daily aggregated ED presentation data was obtained from the ACT Health Directorate and Canberra Health Services [37], which oversees all emergency departments in the territory. The DHR dataset contains approximately 1.6 million episodes involving 535,474 distinct patients [8]. It includes demographic information, such as gender and birth date areas, as well as primary and secondary diagnoses classified under the International Classification of Diseases, tenth revision (ICD-10). Additional variables include admission and discharge dates, triage levels, and patient outcomes, providing a rich resource for investigating trends in emergency care and patient management over time. The details of these variables are listed in Tables A.1 and A.2.

### *2.2. Exogenous variables*

The climate variables used in this study were sourced from the Australian Bureau of Meteorology [38], with readings recorded every three hours. These variables include daily minimum, maximum, and average temperatures (in degrees Celsius), daily mean precipitation (in millimetres), maximum wind gust speed (in kilometres per hour) and its direction (in degrees), as well as the average daily wind speed and wind direction recorded at four distinct time offsets. The data were collected from a weather station in Tuggeranong (Isabella Plains) AWS, the closest meteorological station to Canberra Hospital, covering the period from 1996 to 2022. The ranges and specifics of these climate variables are summarised in Table A.1.

Furthermore, we examine various time-related factors concerning the dates when patients arrive at the ED, including Year, Month, Day of the month, and the day of the week. This is primarily due to the fact that these temporal factors reflect annual and seasonal patterns that influence ED visitation trends. For instance, specific periods of the year, like flu season or festive holidays, often witness a surge in ED attendance. Moreover, temporal patterns can unveil both long-term and short-term fluctuations in ED utilisation, which is crucial for effective resource management.

### *2.3. Predictive Modeling of ED Visitors*

In this study, we developed an extensive ML-based framework to provide an accurate prediction for daily ED visitors' crowdedness. As the nature of the ED data is a time series, we started the implementation using popular sequential models [39], including long short-term memory (LSTM), bi-directional LSTM (Bi-LSTM), and GRU. Two versions of these sequential models were developed: light (one learning layer) and stacked model (two and five learning layers). This is mainly because testing both a single sequential layer and multiple layers in a prediction task is essential for understanding the trade-offs between model complexity, accuracy, and generalisation abilities. As LSTM and GRU are not effective in extracting spatial patterns ( multivariate time series with spatial relationships), we added a convolutional layer into LSTM and GRU models to make a hybrid Convolutional LSTM (ConvLSTM) [40] or hybrid CNN-LSTM architecture.

Other well-known ML-based models for forecasting the ED visitors number are Neural networks such as MLP, Dense neural networks (DNN), etc. The primary reason for applying NN models is their ability to model complex, non-linear relationships in data [41]. In the following, Decision trees (DT), Random Forest (RF), and Extra Trees (ExT) models are commonly used in predicting ED visitor numbers due to their flexibility, interpretability, and ability to model complex relationships among the features. Since our dataset is a combination of various features, demographics, climate, Diagnosis and calendar and makes heterogeneous data, we implemented and tested the effectiveness of six Ensemble models [42], including Adaboost, XGBoost, LightBoost, CatBoost, Gradient Boosting Regression (GBR), and Histogram-based Gradient Boosting Regression (HGBR). The configurations of the ML-based models applied in this study are listed in Table 1.

#### *2.3.1. Meta learning models*

Meta ensemble learning modelling, a sophisticated form of stacking ML methodology, is designed to foster fit diversity by amalgamating predictions from a variety of methods [50]. This approach involves training a meta-learner, often considered the second-level learner, to amalgamate predictions from first-level learners, also known as base models. By integrating the strengths of diverse base models (sub-learners), meta-ensemble models aim to enhance predictive accuracy and robustness [51]. While the conventional stacking technique employs a two-level hierarchy with level-0 and level-1 models, the methodology can be extended to include multiple layers, incorporating various level-1 models and a single level-2 model for improved prediction amalgamation. The core premise of stacking lies in leveraging the distinct strengths of individual models to mitigate weaknesses, culminating in superior performance compared to standalone models.

One pivotal motivation behind stacking is the recognition that each base model possesses unique strengths and limitations [52]. By harnessing the collective power of these models, stacking optimally exploits their strengths while compensating for individual weaknesses, resulting in superior predictive performance. Besides, stacking serves as a potent tool in combating overfitting, a prevalent challenge in machine learning. By training multiple base models on diverse subsets of data or utilising distinct algorithms, the risk of overfitting the training dataset is significantly reduced. The amalgamation of predictions from these diverse models not only enhances generalisation to unseen data but also bolsters the model's overall robustness.

Moreover, the versatility afforded by stacking extends to enabling a flexible and tailored approach to model selection [5]. Rather than relying on a single model, practitioners can leverage a spectrum of base models tailored to address specific facets of the problem at hand. For instance, a random forest model may capture non-linear relationships, a linear regression model can elucidate linear dependencies, and an MLP can unravel intricate feature interactions. Through the synergistic integration of predictions from these specialised models, stacking facilitates the creation of a holistic and refined predictive model that excels in capturing the complexities inherent in diverse datasets.

Table 1: The technical settings of the Machine and Deep learning methods

| # | Acronym | Full name | Hyper-parameters |
|---|---|---|---|
| 1 | LSTM, BiLSTM | Long short-term memory | Hidden Units=32, activation='relu', batch size= 256, dropout=0.2, epochs=200, learning rate=1.0$e-4$ |
| 2 | GRU | Gated recurrent units | Hidden Units=32, activation=relu, batch size= 256, dropout=0.2, epochs=200, learning rate=1.0$e-4$ |
| 3 | CNN | Convolutional neural networks | filter Size=7, num Filters= [64, 32, 16], activation=relu, batch size= 256, dropout=0.2, epochs=300, learning rate=1.0$e-4$ |
| 4 | RF | Random Forest | number estimators=100, max depth=None, min samples split=2, min samples leaf=1, max features=sqrt, |
| 5 | GBR [43] | Gradient Boosting Regression | loss=squared error, learning rate=0.1, number estimators=100, subsample=1.0, min samples split=2, min samples leaf=1, max depth=3, $\alpha = 0.9$ |
| 6 | ExT | Extra Tree | number estimators=100, criterion=gini, max depth=None, min samples split=2, min samples leaf=1 |
| 7 | DNN [44] | Dense Neural networks | Neuron number=32, 16, and 8, kernel initialiser=normal, activation=relu, lr = 0.0001, Optimiser=Adam |
| 8 | DT | Decision Tree Regressor | criterion=squared error, splitter=best, max depth= $D$, min samples split=2, min samples leaf=1, min weight fraction leaf=0.0, |
| 9 | MLP | Multi-layer Perceptron | solver=adam, activation=relu, alpha=1e-4, hidden layer sizes=(200,20,), maxiter=1000 |
| 10 | HGBR [45] | Histogram-based Gradient Boosting Regression | loss=squared error, learning rate=0.1, max iter=100, max-leaf nodes=31, max depth=None, min samples leaf=20, max bins=255 |
| 11 | AdaB [46] | Adaptive Boosting | number estimators=50, learning rate=1.0, loss=linear, base estimator=deprecated |
| 12 | XGB [47, 43] | Extreme Gradient Boosting | Number of estimators=10,max depth=10 , $\gamma = 2$, $\eta = 0.99$, $reg_\alpha = 0.5$, $reg_\lambda = 0.5$ |
| 13 | LGBM [48] | LightBoost | metric='rmse', num iterations=50, num leaves= 100, learning rate= 0.001, feature fraction= 0.9, max depth= 10 |
| 14 | CatB [49] | Categorical Boosting | learning rate=0.01, iterations=1200 |

The multi-layered meta-ensemble modelling paradigm not only amplifies predictive accuracy and robustness but also serves as a versatile and effective strategy for model selection, offering a comprehensive approach to addressing the intricacies of machine learning tasks.

### *2.3.2. Categorical Boosting Method: CatBoost*

The gradient boosting framework is used to build CatBoost [49], which incrementally constructs an ensemble of decision trees. Each previous tree's mistakes are corrected by each new tree, thus forming a highly accurate predictive model. Binary decision trees are used as the base models in CatBoost. These trees recursively partition the feature space $\mathbb{R}^m$ into distinct, non-overlapping regions (or nodes) through binary splits based on splitting attributes $a$. A binary variable generally checks whether a feature $x^k$ surpasses a certain threshold $t$. Formally, this is represented as $a = 1_{x^k > t}$, where $x^k$ can be either a numerical or a binary feature, and $t$ is set to 0.5 for binary features. The decision tree assigns a value to each terminal node (leaf) to estimate the response variable $y$ for regression tasks, predict the class label for classification tasks, or estimate the target for regression problems. Thus, inputs are mapped to predicted outputs by a decision tree $h(\mathbf{x}) = \sum_{j=1}^{J} b_j \prod_{\{\mathbf{x} \in R_j\}}$ based on the following splits defined by these attributes, the disjoint regions corresponding to the tree's leaves are shown by $R_j$.

Two boosting modes are offered by CatBoost: Ordered and Plain. The standard Gradient Boosting Decision Tree (GBDT) algorithm is followed by the Plain mode, but it includes an inbuilt ordered target statistic (TS) handling. In the Ordered boosting mode, supporting models $M_{r,j}$ are maintained during the learning process, where $M_{r,j}(i)$ represents the current prediction for the $i$-th example based on the first $j$ examples in the permutation $\sigma_r$. At each iteration $t$ of the algorithm, a random permutation $\sigma_r$ is sampled from the set $\sigma_1, \ldots, \sigma_s$, and a tree $T_t$ is constructed based on this permutation. First, all target statistics (TS) for categorical features are computed according to the selected permutation. Then, the permutation influences the procedure for learning the tree. When building a tree in CatBoost, two possible scenarios can occur. In the first case, if the model operates under the plain boosting mode, the process involves evaluating candidate splits. The leaf value $\Delta(i)$ for a given example $i$ is calculated by taking the average of the gradients $\text{grad}_{r,\sigma_r(i)-1}$ from the preceding examples $p$ that reside in the same leaf leaf $f_r(i)$ as the example $i$. In this scenario, the split evaluation and leaf value computation are determined based on the gradients of prior examples within the same leaf, as shown in Equation (1).

$$\Delta(i) \leftarrow \text{avg}\,(\text{grad}_r\ (p) \text{ for },p\text{: leaf}_r\ (p) = \text{leaf}_r\ (i)) \text{ for } i = 1 \ldots n; \quad (1)$$

Also, if the mode was equal to the ordered state, Equation 2 will be followed.

$$\Delta(i) \leftarrow \text{avg}\left(\text{grad}_{r,\sigma_r(i)-1}\ (p)\right) \text{ for } p\text{: leaf}_r\ (p) = \text{leaf}_r\ (i), \sigma_r(p) < \sigma_r(i)) \quad (2)$$

*2.3.3. Light Gradient Boosting Machine: LightGBM*

The Light Gradient Boosting Machine, often referred to as LightGBM [48], represents a sophisticated evolution of the GBDT. It enhances the conventional decision tree algorithm by incorporating two innovative techniques that are quite groundbreaking: gradient-based one-side sampling, which is more widely known as GOSS, and a strategy called exclusive feature bundling, abbreviated as EFB. Although the traditional GBDT method is capable of transforming continuous features into discrete ones, it predominantly relies on first-order derivatives to effectively minimise the loss function during the learning process. Furthermore, the decision trees constructed within the GBDT framework are limited in scope to regression trees, as each individual tree is meticulously trained based on the residuals and outputs produced by the preceding trees in the sequence. However, as the size of the dataset escalates, GBDT often encounters significant challenges with both accuracy and efficiency, which can hinder its performance. LightGBM effectively mitigates these limitations by utilising a histogram-based approach to decision tree construction in conjunction with a leaf-wise growth strategy that comes with depth constraints to ensure balanced growth. This combination of features culminates in superior overall performance outcomes, including a reduced incidence of false positive rates and a decrease in the number of missed detections during classification tasks. When provided with a supervised dataset represented as $X$ = $\{(xi,yi)\}_{i}^{N}{}_{=1}$, LightGBM is meticulously designed to minimise a particular regularised objective function that encapsulates the goals of the learning algorithm as follows.

$$Obj = \sum_i l(y_t, \bar{y}_i) + \sum_k \Omega(f_k) \quad (3)$$

The logistic loss function measures the variance between the prediction $\widehat{y_1}$ and the target $y_1$. The regression tree was then used in LightGBM and computed as $F_T(X) = \sum_{t=1}^{T} f_t(x)$.

$$l(\mathrm{y_i}, \hat{\mathrm{y}}_\mathrm{i}) = y_i \ln\left(1 + e^{-\hat{y}_i}\right) + (1 - y_i) \ln\left(1 + e^{\hat{y}_i}\right) \tag{4}$$

The regression tree can alternatively be expressed as $w_{q(x)}$, where $q \in 1,2, \ldots, J$, with $J$ representing the number of leaf nodes in the tree, $q$ denoting the tree's decision rule, and $w$ corresponding to the sample weight. The objective function, in this case, can be written as:

$$Obj^{(t)} = \sum_{\mathrm{i=1}}^{\mathrm{n}} l\left(\mathrm{y_i}, \mathrm{f_{t-1}}(\mathrm{x_i}) + \mathrm{f_t}(\mathrm{x_i})\right) + \sum_k \Omega(f_k) \tag{5}$$

The conventional approach to GBDT employs the steepest descent technique, which fundamentally focuses solely on the gradient derived from the loss function, thereby limiting its consideration to just one aspect of the optimisation landscape. In contrast, the innovative LightGBM algorithm harnesses Newton's method's power, allowing for a rapid and efficient approximation of the objective function, thereby enhancing computational performance and accuracy. Following a meticulous process of simplification and the derivation of Equation 5, it becomes possible to represent the objective function in a more intelligible format as outlined in Equation 6, thus paving the way for a clearer understanding of the underlying mechanics.

$$Obj^{(t)} \cong \sum_{i=1}^{n} \left[g_i f_t(x_i) + \frac{1}{2} h_i f_t^2(x_i)\right] + \sum_k \Omega(f_k) \tag{6}$$

LightGBM offers significant advantages in efficiently and speedly handling large, complex datasets [53]. Its histogrambased decision tree algorithm, multi-threading, and leaf-wise growth strategy enhance training speed, reduce memory usage, and improve accuracy, especially with high-dimensional and sparse data [54]. Techniques like GOSS and EFB maintain accuracy without sacrificing performance, making LightGBM suitable for large-scale ML tasks. However, LightGBM can overfit smaller datasets due to its leaf-wise growth strategy and is more sensitive to hyper-parameter tuning. It may also require more feature engineering for categorical data and needs more interpretability of simpler models. Furthermore, its GPU acceleration is less robust than that of algorithms like XGBoost.

### ***2.4. Proposed Meta-learning Gradient Boosting model***

#### ***2.4.1. Problem formalisation***

In the context of a dataset that encompasses various examples related to Emergency Department (ED) attendance, the primary learning task at hand involves the intricate process of constructing predictive models aimed at forecasting future attendance patterns with the utmost precision and reliability. In a standard ED forecasting scenario that occurs at a specific time denoted as $t$, a comprehensive set of input variables represented as $\vec{A}(t) = [a_1(t), \ldots, a_N(t)]$ is meticulously provided, which is then paired with the corresponding output or target variable $z(t)$, representing the number of ED attendances, thereby forming a cohesive pair denoted as $\langle \vec{A}(t), z(t) \rangle$. For the analytical purposes of this study, time is conveniently expressed in weekly intervals, although it is important to note that alternative time units such as days or months could also be seamlessly applied without disrupting the methodology. The examples utilized in this research are systematically structured as a time series, which can be denoted as $\langle \vec{A}(t_0), z(t_0) \rangle, \ldots, \langle \vec{A}(t_L), z(t_L) \rangle$, commencing from the initial time point $t_0$ and extending through to the final time point $t_L$.

The input variables denoted as $a_i(t)$, where $i$ ranges from 1 to $N$, represent a variety of exogenous factors that influence ED attendance patterns. Within the scope of this study, these input variables encompass a diverse range of elements, including climatic conditions, demographic characteristics, ICD-10 diagnosis codes, and various calendar features, as outlined in Tables A.1, and A.2. The calendar features in question consist of nominal variables that serve to identify specific temporal attributes such as the year, month, day of the week, day of the month, and the corresponding season. On the other hand, the climate data is composed of numerical variables that encapsulate daily averages of critical factors such as temperature, humidity levels, wind speed, and wind direction, all of which play a crucial role in influencing health outcomes. Additionally, the demographic information incorporated into the analysis comprises numerical variables that are related to the characteristics of individuals visiting the ED, including vital statistics such as age, gender, and race, which collectively contribute to a comprehensive understanding of attendance trends in the emergency healthcare setting.

Given that ED attendance can be quantified numerically, it becomes evident that the model undergoing training can be expressed as a function denoted by $f(\cdot)$, which is specifically designed to predict the output variable $z(t)$ at a future moment in time $t$ (representing the ED attendances at that next day $t$, where $t$ exceeds $t_L$ ). This predictive output is defined as $\hat{z}(t) = f(\cdot)$. The inputs that feed into this function consist not only of the current array of variables represented as $a_1(t), \dots, a_N(t)$ but also include historical ED attendance figures $z(t - d_j)$, which are associated with various delays $d_j$, where the index $j$ falls within the range of $[1, \dots, S]$, and importantly, $d_S$ remains less than $t_L$. It is assumed that these previous attendance values are readily available and known when forecasting future attendance for the specified time $t$. The overarching framework that guides this prediction process is encapsulated in the formulation provided in Equation 7 , which serves as a foundational element of the model.

$$\hat{z}(t) = f\big(a_1(t), a_2(t), \dots, a_N(t), z(t - d_1), z(t - d_2), \dots, z(t - d_S)\big) \quad (7)$$

The specific modelling technique significantly influences the intricacy of defining $f(\cdot)$. In this particular scenario, we delve into the realm of Meta-learning, which is characterised as a supervised learning approach that trains the function through the utilisation of historical data, comprising a sequence of examples that are structured as $\langle \vec{A}(t_0), z(t_0) \rangle, \dots, \langle \vec{A}(t_L), z(t_L) \rangle$. The primary goal of this process is to minimise the difference that exists between the actual observed values $z(t)$ and the predicted values $\hat{z}(t)$ derived from the known examples that have been used in the training phase. A pivotal challenge that Meta-learning addresses is the risk of overfitting, a condition wherein the model $f(\cdot)$ may become excessively attuned to the particular training dataset, resulting in subpar performance when confronted with new and previously unseen cases. To combat this issue, the methodology incorporates various strategies aimed at ensuring that the model maintains a strong capacity for generalisation when faced with novel scenarios, thereby achieving an effective balance between flexibility and robustness that is essential for reliable predictions.

#### *2.4.2. Meta-learning Gradient Boosting*

To develop a robust and effective predictive model for forecasting ED visitor numbers, we propose a novel meta learning gradient-boosting approach to reduce bias and variance in prediction errors. In the first phase, we designed a comprehensive comparative framework to identify the best sub-learners, evaluating a total of 23 different methods, including ML techniques, sequential deep learning models, and ensemble models. Statistical analysis of the results demonstrated that (See Tables 4 and 5) CatBoost and LightGBM significantly outperformed other models in predicting ED attendance. In the second phase, we selected a diverse set of predictive models to enhance the variety of learning mechanisms, thereby improving the overall performance of the meta-learning ensemble. Notably, the inclusion of Random Forest and Extra Trees models contributed to a marked improvement in prediction accuracy, highlighting the importance of model diversity in reducing error and enhancing robustness.

After selecting the sub-models, our next focus was on identifying the best-performing main learner. Choosing an effective meta-learner is crucial, as it synthesizes the outputs of the base learners into a final prediction. To ensure diversity and robustness, we evaluated several master-learners, including linear regression, MLP, SVM, AdaBoost, and XGBoost. Among these, MLP consistently outperformed the other algorithms in our experiments. As the main learner, MLP's role is to learn the optimal method for weighting the predictions from the base learners. It captures the relationship between the base learners' outputs and the target variable, uncovering patterns and dependencies that enhance predictive accuracy. Through the use of Backpropagation, the standard algorithm for training MLP, it fine-tunes the combination of base learners' predictions, resulting in a more accurate and robust final prediction. This highlights MLP's effectiveness in meta-learning frameworks for improving overall model performance. After applying Recursive Feature Elimination (RFE) to identify the optimal sub-features [55], we developed a fast and efficient hyper-parameter optimization method. This method combines Differential Evolution [56] (DE) with the Nelder-Mead (NM) algorithm, providing a robust approach for fine-tuning model parameters (See Section 2.6). By employing this adaptive technique, we successfully optimized the hyper-parameters for four sub-learners and the Multi-layer Perceptron (MLP). This hybrid strategy enhances the search process for optimal hyper-parameters, ensuring improved model performance while maintaining computational efficiency, especially in complex and high-dimensional feature spaces. The technical details of the proposed Meta-learning model can be seen in Figure 1.

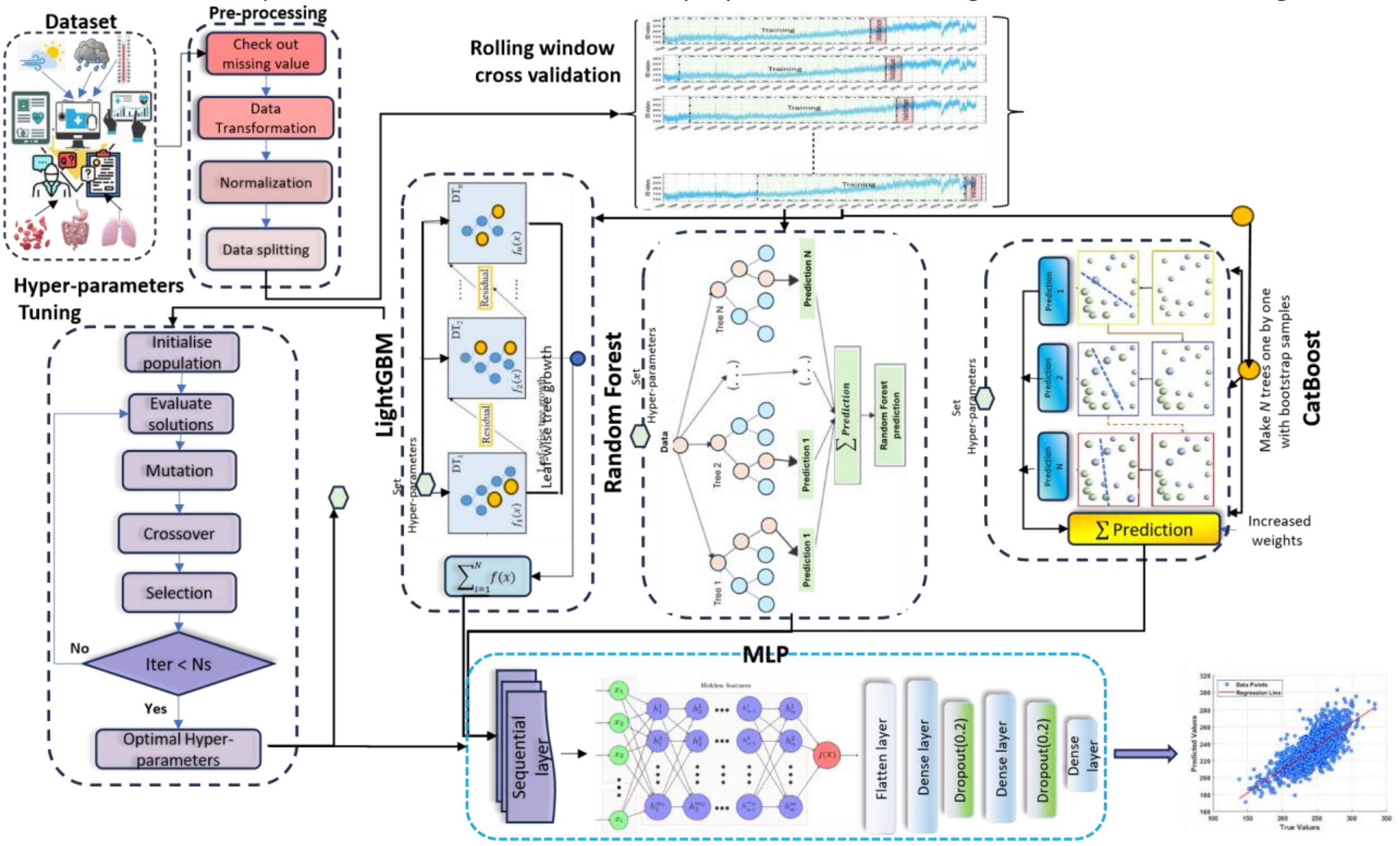

**Figure 1: The landscape of the proposed Meta-learning ensemble architecture.**

### *2.5. Feature Selection*

To optimise the selection of sub-features and enhance the performance of the proposed Meta-learner model, we employed the Recursive Feature Elimination (RFE) [57] technique. RFE is a highly effective method for feature selection, systematically identifying and removing irrelevant or redundant features. By iteratively eliminating the least important features, RFE simplifies the model while maintaining or improving predictive accuracy. This reduction in feature space helps mitigate overfitting, ensuring the model generalises unseen data better. Additionally, RFE enhances model interpretability by focusing on the most significant predictors. Its flexibility and compatibility with a wide range of machine learning algorithms further reinforce its utility in building robust and efficient predictive models. In order to implement RFE, we used Scikit-learn (an open-source ML library), which was developed based on Python and RFE class from [58].

### *2.6. Differential Evolution Hyper-parameters Tuning*

Adjusting hyper-parameters in machine learning is incredibly crucial due to its direct impact on the model's efficacy and capacity to generalise. The effects of hyper-parameter adjustment are deep and varied, as they influence how effectively a model can learn from data and generate predictions. This optimisation endeavour seeks to boost the model's precision, lower error rates, and enhance pertinent metrics, ultimately culminating in superior predictive capabilities. The skilful tuning of hyper-parameters allows the model to adeptly discern the fundamental patterns within the data, thus amplifying its overall effectiveness. Furthermore, hyper-parameter adjustment guarantees that a machine learning model can generalise effectively to new data. Generalisation is vital to model performance, as it dictates how proficiently a model can forecast fresh, unseen instances. One can achieve the optimal equilibrium between underfitting and overfitting by fine-tuning hyper-parameters.

In this research, we demonstrated the adaptability of the XGBoost model by utilisinga powerful optimisations technique, Differential Evolution [59] (DE) alongside Nelder-Mead [60] (NM), to meticulously adjust four hyperparameters. These parameters, including tree depth, count of estimators, learning rate, and subsample rate, were fine-tuned using both accuracy and AUC as evaluative metrics, showcasing the model's outstanding performance. DE represents an innovative approach within evolutionary algorithms, integrating differential vectors into a triangular search pattern and has emerged as a highly esteemed population-based optimisation strategy, frequently leveraged to address a broad spectrum of real-world challenges marked by noise, volatility, and multimodality. Among DE's various facets, the mutation operator is a pivotal component significantly contributing to the algorithm's effectiveness. Numerous mutation operators have been proposed in the DE landscape, each presenting unique convergence and exploration potential rates. A notable mutation strategy, $DE/best/1/bin$, is celebrated for its rapid convergence, especially in the context of unimodal challenges. Yet, its performance tends to falter when confronting the hurdles of local optima, often leading to premature convergence as it navigates through the complexities of multi-modal problem environments. Equation 8 articulates this mutation scheme mathematically, encapsulating its fundamental operational characteristics within DE.

$$\text{DE/ best /1/ bin}: \vec{T}_{k,g} = \vec{S}_{\text{best},g} + \zeta \cdot \left(\vec{S}_{r_1,g} - \vec{S}_{r_2,g}\right) \quad (8)$$

where $\vec{T}_{k,g}$ represents a vector that captures the nuances of three distinct solutions, namely $\vec{S}_{\text{best}}$, $\vec{S}_{r_1}$, and $\vec{S}_{r_2}$. Simultaneously, $\zeta$ acts as the mutation factor, fine-tuning the exploratory capabilities' magnitude and pace. Transitioning to another crucial evolutionary component in DE, we unearth the significant influence of crossover. The primary technique utilized for crossover is the binomial method, defined by a structure that incorporates the trial vector, referred to as $\vec{U}$, alongside the crossover rate, indicated as $C_r$, which varies between zero and one as shown.

$$\vec{U}_{k,j}^{g} = \begin{cases} \vec{T}_{k,j}^{g}, & \text{if ( rand } \le C_r) \text{ or } (j = sn), \\ \vec{S}_{k,j}^{g}, & \text{otherwise.} \end{cases} \quad j = 1,2, \dots D \quad (9)$$

A designated quantity of candidates, denoted as *sn*, is chosen in the crossover phase. This leads to the formation of an innovative solution through a balanced amalgamation of the parent and progeny elements, thereby continuing the evolutionary journey of optimisation, as illustrated.

The integration of DE alongside the Nelder-Mead (NM) algorithm provides a remarkable array of benefits by effectively harnessing the strengths inherent in both global and local optimisation methodologies. While DE excels at conducting an extensive exploration of the global search landscape of hyper-parameters, accurately identifying and locating the most promising regions that may yield optimal configurations for the predictor, NM steps in with its remarkable efficiency to carry out the intricate and crucial task of local optimisation, thereby honing in on the solution with precision. The technical details of this adaptive hyper-parameters optimisation (DNO) can be seen in Algorithm 1.

**Algorithm 1** *Differential Nelder − Mead Optimisation (DNO)*

**procedure** DNO ($solution_0 = h_1, h_2, ..., h_{N_h}$)
**Initialisation**
$N_p = 15, \zeta = 0.5, C_r = 0.5, Max_{iter} = 100$ ▹ Initialise DE parameters
$Pop_0 = Gen(solution_i, i \in [1 : N_h])$ ▹ Generate initial population randomly
$fit = Eval(Pop_0)$ ▹ Evaluate initial population
**Differential Evolution**
**for** $i$ in $[1, .., Max_{iter}]$ **do** ▹ Main loop of hyper-parameters tuning
$r_1, r_2, r_3 = randperm(Pop_i)$ & $r_{best} = Best(Pop_i)$
$P^1 = Pop(r_1), P^2 = Pop(r_2), P^3 = Pop(r_3), Gbest = Pop(r_{best})$
**for** $j$ in $[1, .., N_h]$ **do** ▹ Mutation and crossover loop
**if** $c_r < rand()$ **then**
$Tsol_j = Gbest_j + \zeta \times (P^1_j - P^2_j) + \zeta \times (P^3_j - Gbest_j)$
**else**
$Tsol_j = Pop^i_j$
**end if**
$fit_T = Eval(Tsol)$
**if** $fit_T > fit_j$ **then** ▹ Maximise accuracy of Meta-ED model
$fit_j = fit_T, Pop_j = Tsol$
**end if**
**end for**
$\Delta P = Max(Pop_i) - Max(Pop_{i-1})$ ▹ Compute the optimisation progress
**Nelder-Mead**
**if** $\Delta P < \Lambda$ **then** ▹ Run local search
$< Sol_{nm}, fit_{nm} >$=Nelder-mead($Max(Pop_i)$)
$Best_{pop} = Sol_{nm}, Best_{fit} = fit_{nm}$
**end if**
**end for**
$Best_{solution} = Best_{pop_{Max_{iter}}}$
$\langle fit_{best} \rangle = Train(Meta - ED(Best_{solution}))$
**return** $Best_{solution}, fit_{best}$ ▹ Optimal hyper-parameters
**end procedure**

### 2.7. Explainable AI

While it is true that the intricate and sophisticated pipelines associated with AI-based models have the remarkable capability to produce predictions with a striking level of accuracy, there exist two fundamental elements that are frequently overlooked and left unaddressed: the crucial aspects of understanding and explaining the outcomes generated by these advanced models [61]. The initial phase of understanding is fundamentally focused on the meticulous processes involved in training the AI model, alongside the rigorous quality assurance practices that ensure its reliability and effectiveness in various applications. On the other hand, the explaining phase emerges as an absolutely vital component when an ML model is utilised and applied within the unpredictable and often complex realm of real-world scenarios such as health and medical AI-based models [62], where the stakes can be extraordinarily high. This phase is meticulously crafted not only to interpret the intricate mechanisms through which predictions are produced based on real-world data inputs but also to furnish end users with explanations that are easily comprehensible and accessible to the human mind. Such interpretability takes on heightened importance, especially in mission-critical applications, where having a clear understanding of the decision-making processes that underpin the model's predictions becomes an absolute necessity for users and stakeholders alike.

In the context of AI-based models, complex models such as deep learning and Ensemble methods—which include techniques like XGBoost [63]—are frequently categorised as black box models due to their inherently opaque nature and the lack of transparency surrounding their internal workings. The intricate architectures of these advanced models create significant challenges in discerning the specific contributions that individual features make toward the final prediction and elucidating the complicated interactions between these various features. For example, within the realm of deep learning models [64], it presents a daunting challenge to trace the precise pathways through which particular input features exert influence on the resulting output features, leading to a veil of uncertainty that can obscure the understanding of the model's behaviour. The endeavour to develop practical and comprehensible outcomes that are also explainable yields considerable advantages, including enhanced decision-making capabilities and the vital opportunity to uncover potential biases or sources of discrimination that may exist within the model, thereby allowing for these issues to be addressed in a thoughtful and effective manner.

In this study, we developed comprehensive global and local interpretation frameworks to dissect the predictive behaviour of the proposed Meta-learning model. Global interpretation provides a high-level understanding of the model's overall dynamics, shedding light on the relationships and interactions between features and determining their relative contributions to the model's performance [65]. We utilised advanced techniques such as feature importance analysis and partial dependence plots to quantify feature relevance and visualise the interactions between predictors. These methods offer a holistic view of how key variables shape the model's output across the entire dataset. Conversely, local interpretation delves into the rationale behind individual predictions [66], providing case-specific explanations for why the model suggests certain outcomes. For this, we employed the SHAP (SHapley Additive exPlanations) approach [67], a sophisticated and widely used tool for isolating the contribution of each feature to a specific prediction and formulated as follows.

$$\Omega_i(h,a) = \sum_{z' \subseteq a'} \frac{|z'|!\left(N_f - |z'| - 1\right)!}{N_f!}\left[h_a(z') - h_a(z' \setminus i)\right] \quad (11)$$

where $\Omega$ and $h$ are the Shapley value of $i^{th}$ features and black-box model, and $a$ is the sample, $|z'|$ refers to the subset of sample count which is not equal to zero in $z'$. Also, $N_f$ is the number of features which play the role of inputs in the first layer. By leveraging this method, we were able to illuminate the inner workings of the model's decisionmaking process at a granular level, providing actionable insights into its behaviour on a case-by-case basis. The combination of these global and local interpretation methods ensures a well-rounded and transparent understanding of both the general and instance-specific behaviour of the Meta-learning model, enhancing its interpretability and trustworthiness.

## 3. Results and Discussions

### *3.1. Study design and data sources*

The Emergency Department at Canberra Hospital is crucial in providing healthcare services to the ACT populace. This investigation harnesses the DHR dataset sourced from the ED to delve into patient profiles, treatment trends, and outcomes. Spanning an impressive 23 years, from January 1999 through December 2022, the dataset is a treasure trove of insights for in-depth analysis. The DHR dataset comprises around 1.6 million episodes involving 535,474 distinct patients. It includes a demographic registry that details sex and birth dates. Each care event's primary diagnosis is categorised using the International Classification of Diseases, tenth revision (ICD-10), while up to 10 most frequent diagnoses follow ICD-10 coding. Additional information encompasses admission and discharge dates, triage levels, and patient outcomes. Canberra Hospital's Emergency Department processes a considerable influx of patients, with 76,505 reported presentations at ACT public hospitals [68]. In the first and second quarters of 2023-24, 61.8% of patients received timely ED treatment, with an average wait time of 25 minutes. Moreover, 55.6% of patients were released from the ED within four hours of arrival, while 3.9% opted not to wait for assessment [68].

### *3.2. Characteristics of participants*

Figure 2 illustrates a histogram detailing various parameters linked to the number of ED visitors. Considerably, in 2019, the highest surge in ED visitor numbers, exceeding 90,000 episodes, was attributed to the COVID-19 pandemic. Moreover, the peak traffic occurs in August, on Mondays, between 10:00 AM and 1:00 PM, surpassing $1.35 \times 10^5$, $2.35 \times 10^5$, and $9 \times 10^4$ episodes, respectively. In terms of gender distribution, males contributed more significantly than females, with a marginal 2% difference. Statistical analyses further reveal that children (*age* ≤ 14) constitute the predominant demographic, comprising 23% of all ED episodes. Interestingly, the data indicates a negative correlation between visitor age and ED attendance rates. Additionally, concerning triage levels, levels 4 and 3 were the most frequently recorded situations in the ED episodes. These insights gleaned from the histogram shed light on the complex dynamics of ED utilisation and demographics, aiding in the refinement of healthcare resource allocation and patient care strategies.

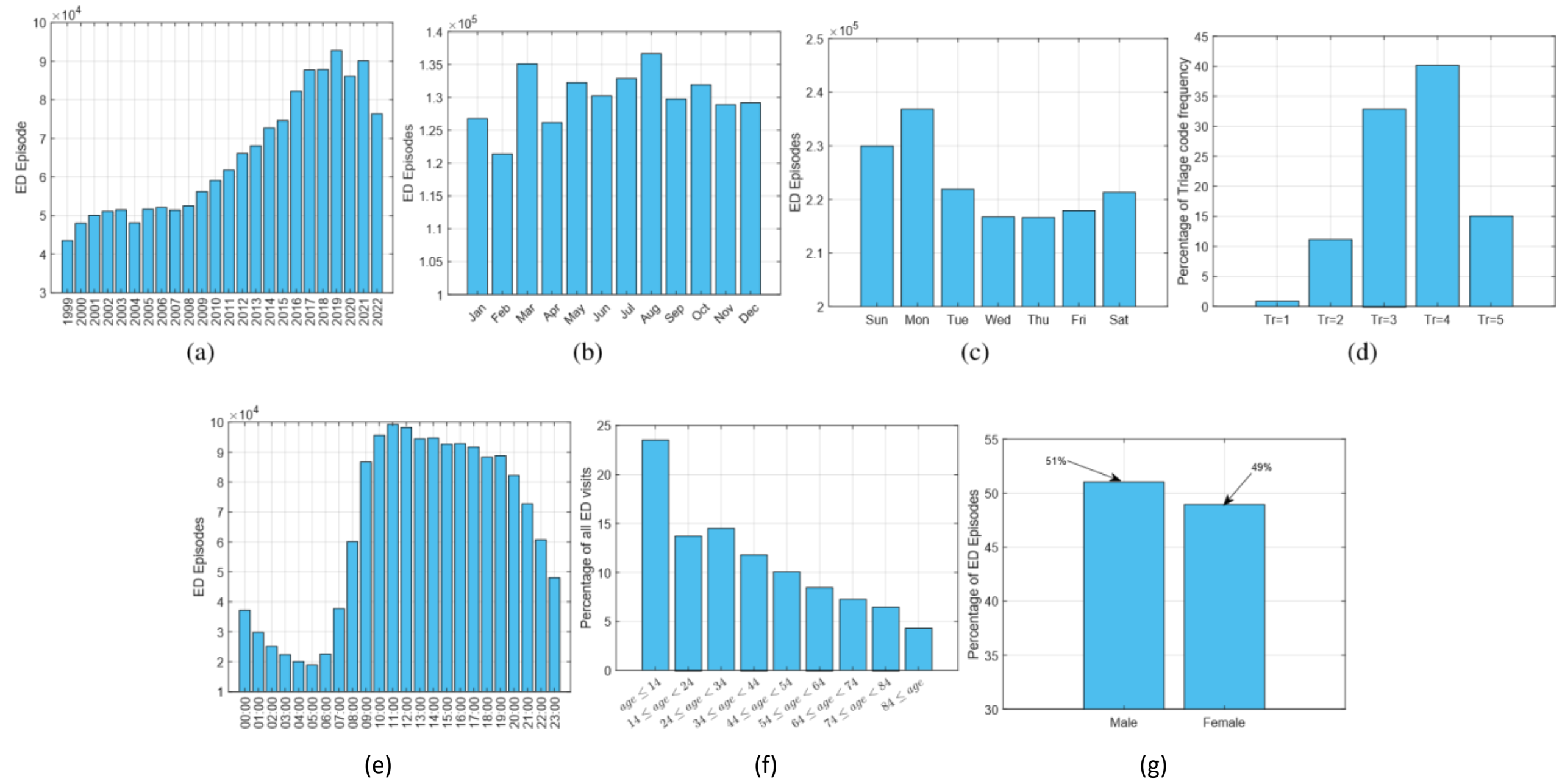

**Figure 2: histogram detailing various parameters linked to the number of ED visitors at Canberra Hospital between 1999 and 2022.**

Table 2 presents descriptive statistics for the ED visitors' population based on 535,474 recorded episodes from 1999 to 2022. The statistics provide insights into various demographic and clinical characteristics of the population served by the Department of Health and Rehabilitation (DHR). Percentages shown in brackets represent values relative to the column base, offering a comparative view of each attribute within the dataset. This analysis aims to highlight trends and patterns in ED usage over the 23-year period.

**Table 2: Descriptive statistics for the visitors' population based on ED episodes (population=535,474) from 1999 to 2022. Percentages in brackets are relative to the column base. Department of Health and Rehabilitation (DHR)**

| Variable | | Total (%) |
|---|---|---|
| Total population | | 535474 (100) |
| Total Episode number | | 1,561,222 (100) |
| Female | | 764,998 (49) |
| Age | <=14 | 366,850 (23.50) |
| | 14-24 | 238,753 (15.29) |
| | 25-34 | 221,889 (14.21) |
| | 35-44 | 180,278 (11.55) |
| | 45-54 | 154,698 (9.91) |
| | 55-64 | 129,775 (8.31) |
| | 65-74 | 112,191 (7.19) |
| | >74 | 156,788 (10.04) |
| Triage | 1 (Red) | 13,237 (0.85) |
| | 2 (Orange) | 174,000 (11.15) |
| | 3 (Yellow) | 513,686 (32.90) |
| | 4 (Green) | 626,124 (40.10) |
| | 5 (Blue) | 234,175 (15.00) |
| State | ACT<br>NSW | 1,315,365 (84.25)<br>218,408 (13.99) |
| Disposition | Admit<br>Home<br>Did not wait (DNW)<br>Left at own risk (LOR) before treatment completed<br>Discharged to DHR<br>Referred to other TCH service<br>Transferred to other hospital<br>Went to GP<br>Died in ED<br>Dead on arrival | 503,784 (32.27)<br>911,854 (58.41)<br>103,379 (6.62)<br>7,181 (0.46)<br>21 (0.0013)<br>25,938 (1.66)<br>6,756 (0.43)<br>1,236 (0.079)<br>1,070 (0.068)<br>3 (0.0001) |
| Most frequent ICD-10 | ’Z53.1’ (Did not wait for treatment)<br>’R07.4’ (Chest pain)<br>’R10.4’ (Pain in abdomen)<br>’B34.9’ (Viral infection)<br>’Z09.9’ (For review)<br>’J45.9’ (Asthma, acute)<br>’N39.0’ (Urinary tract infection)<br>’R11’ (Nausea and vomiting except in pregnancy)<br>’S09.9’ (Injury, unspecified or suspected of head)<br>’M54.5’ (Low back pain)<br>’R45.81’ (Suicidal ideation) | 89,874 (5.76)<br>62,962(4.03)<br>62,511(4.00)<br>31,155 (2.00)<br>30,513 (1.95)<br>19,436 (1.24)<br>18,028 (1.15)<br>16,532 (1.06)<br>15,874 (1.02)<br>14,060(0.90)<br>13,425 (0.86) |

The analysis of the 20 most common reasons for visiting the ED reveals that the group of patients who did not wait for treatment had the highest number of visitors. In the following, Chest pain, Pain in the abdomen and Viral infections are second to fourth rank among all 20 viral ICD-10 recorded (See Figure 3). This high prevalence of chest pain, abdominal pain, and viral infections as leading reasons for ED visits highlights the need for effective triage and resource distribution, as these conditions are common and can require urgent attention, depending on severity.

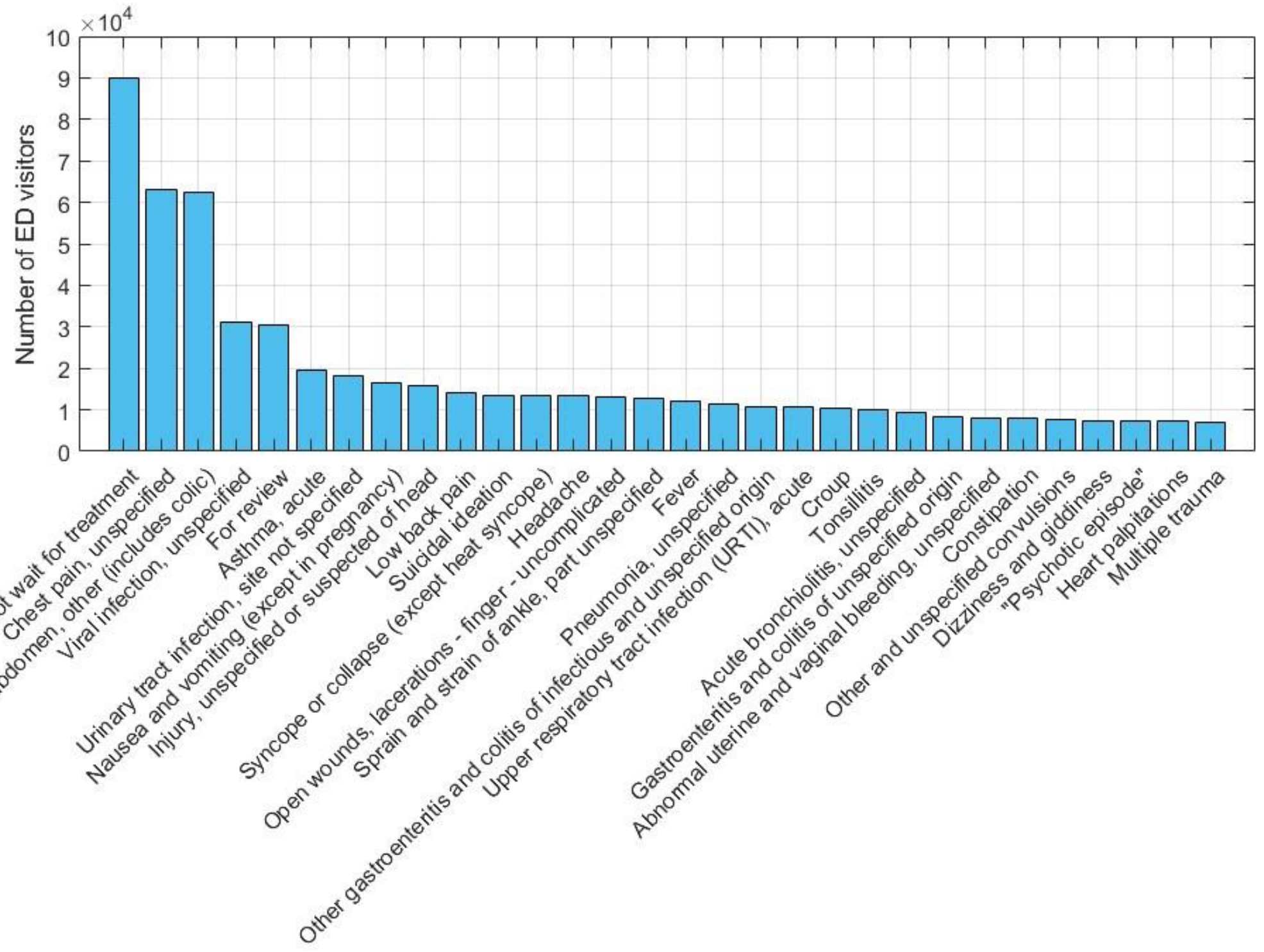

**Figure 3: The distribution of the 30 most prevalent reasons for Emergency Department (ED) visits was determined through an analysis of a population size of 535,474 visits to the Canberra Hospital ED between 1999 and 2022, encompassing a total of 1.6 million episodes. Except for 'Did not wait for treatment', the highest frequencies were observed for 'Chest pain', 'Abdominal pain', and 'Viral infection', highlighting their notable prevalence among ED visitors.**

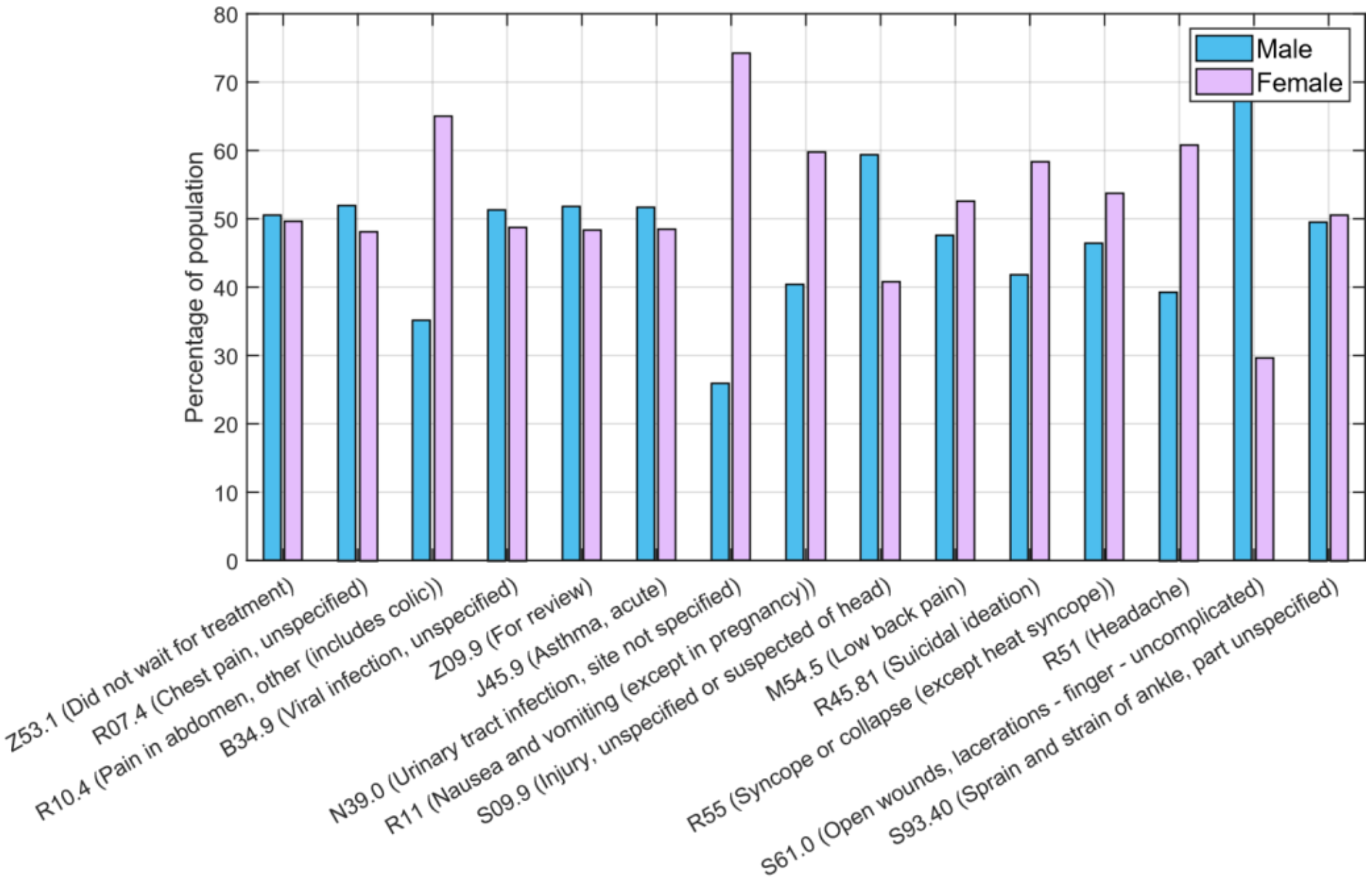

Figure 4: The distribution percentage of the top 15 most viral reasons for ED visitors based on gender.

The insights gleaned from Figure 4, which delineates the distribution percentages of the top 15 reasons for ED visits stratified by gender, reveal intriguing trends. The data indicates that males exhibited a visit rate for open wounds that were approximately double that of females, implying a higher prevalence of this issue among male ED attendees. In contrast, females demonstrated a visit rate for urinary tract infections that was roughly twice that of males, signifying a higher incidence of this ailment among female ED visitors. Moreover, in cases of abdominal pain, nausea, vomiting, lower back pain, suicidal ideation, syncope, and headaches, women displayed a higher frequency of ED visits compared to men, underscoring a distinct gender bias in emergency care utilization for these health concerns. Furthermore, in 7 out of the 15 leading reasons, females outpaced males in their contribution to these visits. This data underscores the importance of tailoring healthcare resources to address the specific needs of women.

The analysis depicted in Figure 5 unveils several significant patterns within the ED visits. Firstly, it is evident that individuals younger than 14 years exhibited the highest incidence of visits associated with Viral infections, Abdominal pain, Asthma, Nausea, vomiting, and injuries. Conversely, seniors aged 64 and above presented the highest frequency of ED visits relating to Chest pain, Urinary tract infections, lower back pain, and Syncope. Furthermore, the younger population, particularly those aged between 14 and 24, predominantly sought medical attention for issues such as Suicidal ideation, Open wounds, finger lacerations, and Ankle sprains and strains. These observations underscore distinct health concerns across different age groups accessing emergency care services.

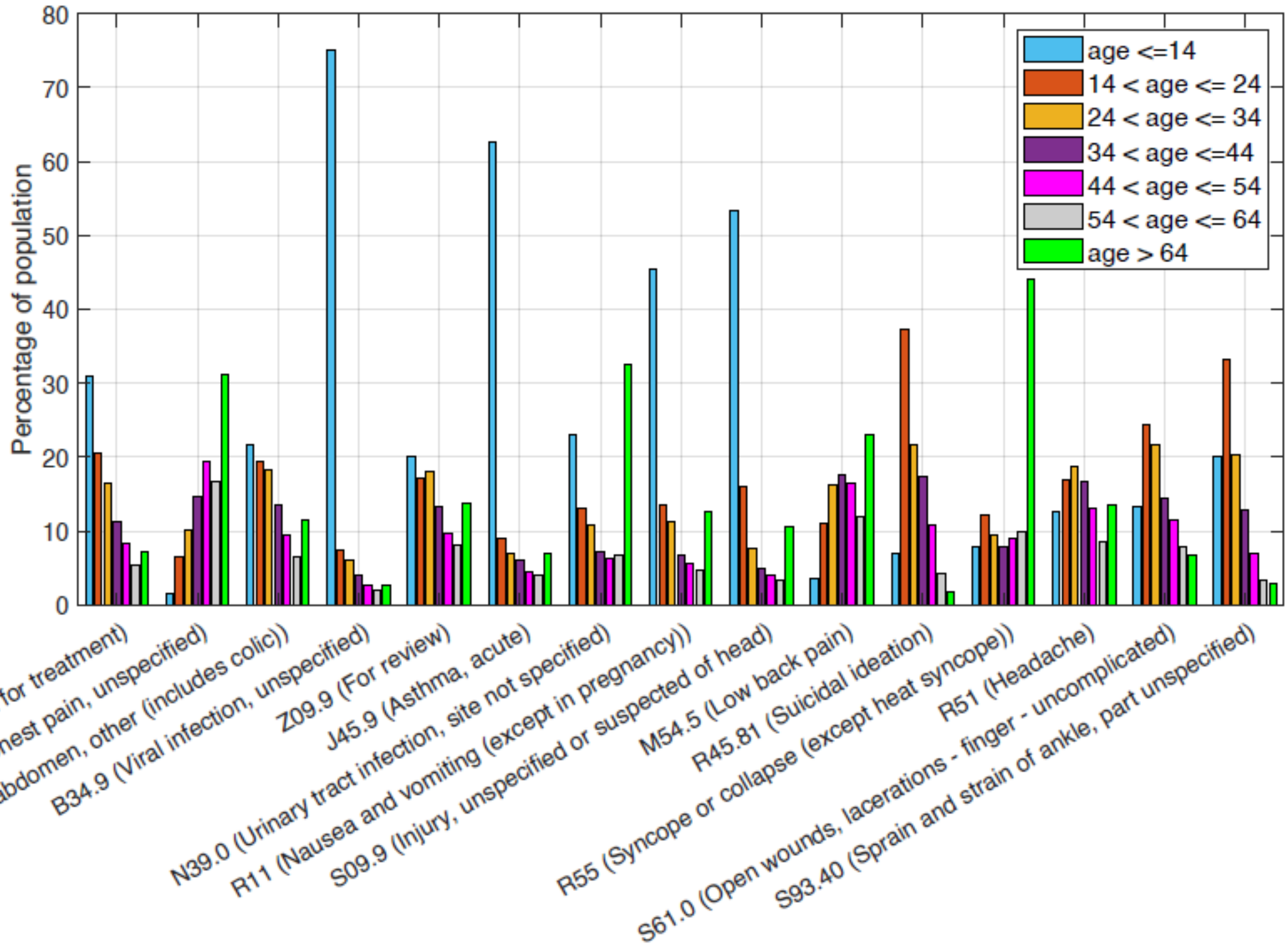

Figure 5: The distribution percentage of the top 15 most viral reasons for ED visitors based on the age groups.

As seen in Figure 6, the feature exhibiting the highest correlation with the number of ED visitors is the admission year, demonstrating a strong correlation coefficient of 88%. Interestingly, the top five most common ICD-10 codes correlated with visitor numbers range from 34% to 72%, with lower-ranked codes showing unexpectedly higher correlation coefficients. Moreover, a distinctive correlation pattern emerges between age groups and ED visitor numbers. At the same time, there is a direct relationship between increasing age and ED attendance for most age groups; individuals under 50 years old exhibit a negative correlation. This suggests that younger individuals may be less inclined to visit the ED due to factors such as better overall health, lower prevalence of chronic conditions, and fewer emergencies. Noteworthy positive correlations are observed among chronic disorders, with Cardiovascular, Mental Behaviours, and Genitourinary disorders showing substantial correlations of 51%, 60%, and 68%, respectively. Furthermore, we can see that out of 13 climate variables, the average wind speed stands out with the highest negative correlation with ED visitors at -28%. In contrast, other weather variables do not exhibit significant correlations in this context.

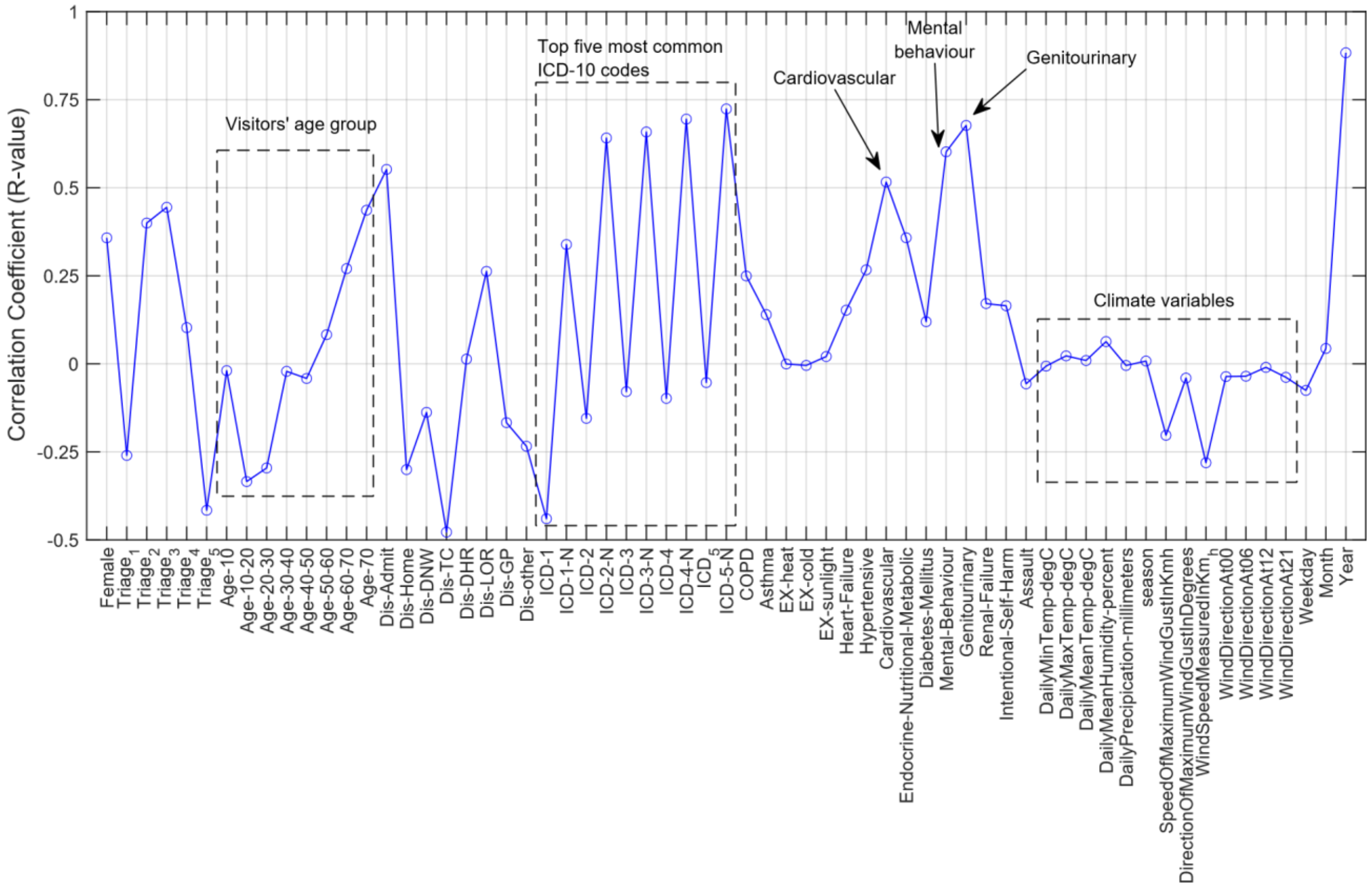

**Figure 6: The correlations between the ED visitors' number and wide range of features, including patient demographic, medical information, admission details and climate variables.**

The correlation analysis (See Figure A.2) reveals fascinating insights regarding the relationship between climate variables and ED visitors. Considerably, the daily temperature exhibits the highest correlation with ED visitor age groups, showcasing a negative correlation of 30% with individuals younger than ten years old and approximately 10% with other age groups. Moreover, an increase in temperature is linked to a rise in genitourinary diagnoses in the ED, indicating a significant correlation of 10%. Humidity levels also play a crucial role, particularly impacting children under ten years old by 15% and increasing the risk of asthma by 10%. Seasonal variations are paramount, with the season highly correlated with ED visitors in the under-10 age group at 20%, emphasising the necessity for tailored seasonal health services for children. Wind gust speed emerges as the most influential climate variable, displaying a notable negative correlation of 20% with the number of ED visitors. This rate suggests that higher wind speeds correspond to decreased ED visits. Additionally, wind speed exhibits negative correlations with mental disorders and genitourinary risks, underscoring the multifaceted impact of climatic factors on health outcomes in the ED setting.

In order to extend the analysis related to climate variables, Figure 7 depicts the relationship between climate variables and the top ten most common diagnoses based on ICD-10 codes at Canberra Hospital's Emergency Department. Several notable observations emerge from this analysis: Viral infection exhibits a substantial positive correlation with both seasonal patterns and humidity levels, standing at 17% and 10%, respectively. Conversely, viral infection negatively correlates with temperature and wind speed, exceeding -20% and -15%, respectively. This observation implies that patients are likely to be admitted with viral infections as temperatures decrease. In the case of chest pain, a 13% correlation is observed with rising humidity levels. Interestingly, the risk of chest pain disorders shows an 18% increase when wind speed decreases.

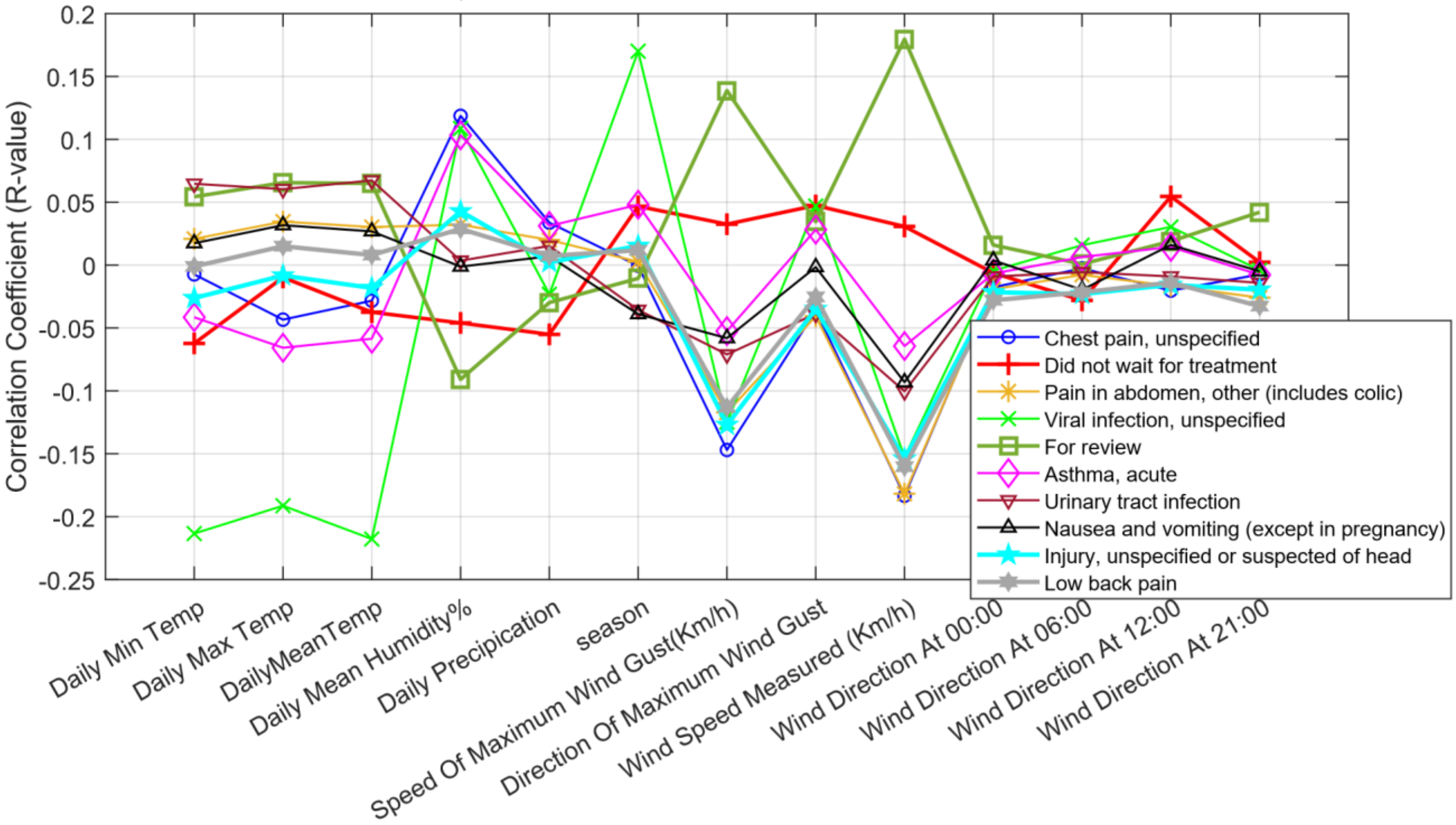

Figure 7: The correlations between climate variables and top ten most common diagnoses based on ICD-10 code.

The analysis presented in Figure 8 unveils a convoluted and non-linear association between climate parameters and the volume of ED visitors. Notably, Figure 8 (c) illustrates a noteworthy trend where elevating wind speeds (Km/h) correspond to a reduction in the influx of individuals seeking care at the ED. This unexpected inverse correlation suggests that heightened wind conditions may have a mitigating effect, leading to a decreased demand for emergency medical services during such circumstances. Conversely, despite particular studies, a substantial relationship between temperature variations and ED visitation rates remains elusive in the data.

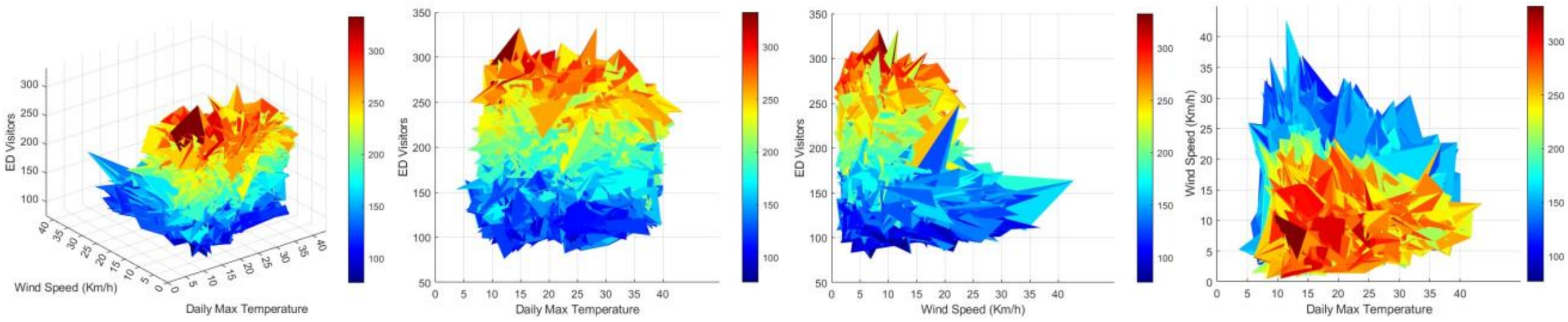

(a) (b) (c) (d)

**Figure 8: (a) A 3D correlation between number of ED visitors, wind speed (Km/h), and daily maximum temperature. (b) correlation between ED visitors number and daily maximum temperature. (c) correlation between ED visitors number and wind speed. (d) correlation between daily maximum temperature and wind speed.**

A comprehensive analysis of the correlation between wind directions and ED visitor numbers has revealed distinct patterns. Specifically, findings indicate a notable surge in the volume of ED visitors when winds blow from the South and Southeast directions. This observed relationship suggests a potential influence of these wind patterns on health outcomes within the community, potentially leading to higher incidents of health-related emergencies (See Figure A.3).

### 3.3. Setup of implementations

A comprehensive study explored a diverse array of machine learning models encompassing traditional and deep learning architectures. Sequential models like LSTM, GRU, BiLSTM, and various stacked models, alongside Convolutional Neural Networks (CNN) and hybrid CNNs, were meticulously implemented using MATLAB (R2024a) in conjunction with the Deep Learning Toolbox. These models were tailored to handle sequential data, time series analysis, and image processing tasks with a focus on accuracy and efficiency. A parallel suite of machine learning algorithms, including MLP, Deep Neural Networks (DNN), Random Forest, tree-based models, ensemble methods, and a meta-learning framework were expertly crafted utilising Python-based libraries. Specifically, Spyder (4.1.5) served as the integrated development environment within the Anaconda (2.1.1) Python distribution (3.7.9), leveraging the power of Keras (2.12) and scikit-learn (1.2.2) for model development. Keras, built on top of TensorFlow (2.12.0), a cornerstone of modern machine learning infrastructures developed by Google, facilitated seamless model creation and training. To bolster the meta-learning framework, boosting libraries such as XGBoost, Catboost, LightGBM, and HGBR were judiciously employed, ensuring robust and versatile meta-learner capabilities that could adapt to a wide range of datasets with precision and agility.

### 3.4. Performance evaluation metrics

In this section, we emphasise the critical importance of selecting and utilizing a diverse set of evaluation metrics when assessing the effectiveness of regression prediction models. To ensure a robust validation process on a reserved test set and mitigate risks of overfitting or underfitting, careful consideration of metrics aligned with task objectives and dataset characteristics is essential. This study incorporates two key groups of metrics—accuracy and bias—to provide a comprehensive assessment framework. While accuracy is fundamental, it can be misleading in cases of imbalanced data distributions. Additional learning error metrics are crucial for identifying systematic prediction deviations that may stem from data representation issues, model limitations, or flaws in training methodology. Bias can lead to persistent mispredictions, distorting overall performance evaluations. A detailed overview of 10 evaluation metrics used in the study is presented in Table 3 to aid readers in understanding the nuances of the evaluation process and aside from the metrics such as the Pearson correlation coefficient, Explained variance score, and Mean Directional Accuracy, the remaining metrics should be minimised.

**Table 3: The ML and deep learning evaluation metrics.**

| Metrics | Equation |
|---|---|
| Pearson correlation coefficient | $R = \frac{\frac{1}{N_s}\sum_{k=1}^{N_s}(f_e(k)-\overline{f}_e)(f_t(k)-\overline{f}_t)}{\sqrt{\frac{1}{N_s}\sum_{k=1}^{N_s}(f_e(k)-\overline{f}_e)^2}\times\sqrt{\frac{1}{N_s}\sum_{k=1}^{N_s}(f_e(k)-\overline{f}_t)^2}}$ |
| Explained variance score | $EVS = 1 - \frac{\sum_{k=1}^{N_s} Variance(f_t(k)-f_e(k))}{Variance(f_t(k))}$ |
| Mean absolute error | $\text{MAE} = \frac{1}{N_s}\sum_{k=1}^{N_s}\lvert f_e(k) - f_t(k)\rvert$ |
| Mean squared log error | $\text{MSLE}=\frac{1}{N_s}\sum_{k=1}^{N_s}(log_e(1 + f_t(k)) - log_e(1 + f_e(k)))^2$ |
| Root mean square error | $\text{RMSE} = \sqrt{\frac{1}{N_s}\sum_{k=1}^{N_s}(f_e(k) - f_t(k))^2}$ |
| Symmetric mean absolute percentage error | $\text{SMAPE}= \frac{1}{N_s}\sum_{k=1}^{N_s}\frac{\lvert f_t(k)-f_e(k)\rvert}{(\frac{1}{2}(f_t(k)+f_e(k)))} \times 100$ |
| Median Absolute Deviation Error | $\text{MEDAE}= \frac{1}{N_s}\sum_{k=1}^{N_s}\lvert f_t(k) - \mu\rvert, \mu = mean$ |
| Mean Absolute Percentage Error | $\text{MAPE}=\frac{1}{N_s}\sum_{k=1}^{N_s}\left\lvert\frac{f_t(k)-f_e(k)}{f_t(k)}\right\rvert$ |
| Mean Directional Accuracy | $\text{MDA}= \frac{1}{N_s}\sum_k \mathbf{1}_{\text{sgn}(f_r(k)-f_r(k-1))=\text{sgn}(f_e(k)-f_r(k-1))}$, $sgn(\cdot)$ and $\mathbf{1}$ are sign and indicator function |
| Relative Square Error | $\text{RSE}= \frac{\sum_{k=1}^{N_s}(f_t(k)-f_e(k))^2}{\sum_{k=1}^{N_s}(\mu-f_t(k))^2}, \mu = mean$ |

### *3.5. Fundamental prediction results*

In this section, we review and compare the performance of ten sequential deep learning models in predicting daily ED visitor numbers and assess the impact of the number of layers on model outcomes. Table 4 presents the detailed statistical prediction results of the ten time-series models, including LSTM, BiLSTM, and GRU. Additionally, we evaluated the stacked architectures of these three models, comparing versions with two layers (denoted as s-LSTM) and larger models with four learning layers (denoted as l-LSTM). The results indicated that BiLSTM achieved the best accuracy (R-value) compared to LSTM and GRU. Furthermore, increasing the number of learning layers resulted in only a marginal improvement in average prediction accuracy (around 2%). Due to the increased training time, this additional complexity is not recommended if the sample size is below 10,000.

**Table 4: Statistical results of ED visitor prediction for standard recurrent networks and small and large multi-layers LSTM, GRU and BiLSTM models and CNN**

| LSTM | | | | | | | | | | BiLSTM | | | | | | | | | |
|---|---|---|---|---|---|---|---|---|---|---|---|---|---|---|---|---|---|---|---|
| Metric | RMSE | MAE | R | MSLE | MEDAE | EVS | MAPE | MDA | RSE | Metric | RMSE | MAE | R | MSLE | MEDAE | EVS | MAPE | MDA | RSE |
| Min | 26.137 | 21.429 | 0.707 | 0.012 | 19.182 | 0.500 | 8.514 | 0.675 | 0.811 | Min | 24.092 | 19.528 | 0.741 | 0.010 | 17.264 | 0.549 | 7.781 | 0.687 | 0.689 |
| Max | 37.354 | 32.528 | 0.772 | 0.026 | 31.770 | 0.595 | 12.914 | 0.696 | 1.657 | Max | 28.949 | 24.124 | 0.781 | 0.015 | 21.794 | 0.609 | 9.582 | 0.709 | 0.995 |
| Mean | 30.571 | 25.773 | 0.741 | 0.017 | 23.990 | 0.547 | 10.230 | 0.683 | 1.124 | Mean | 26.079 | 21.428 | 0.762 | 0.012 | 19.256 | 0.579 | 8.522 | 0.698 | 0.810 |
| STD | 3.7E+00 | 3.6E+00 | 2.3E-02 | 4.6E-03 | 4.1E+00 | 3.5E-02 | 1.4E+00 | 7.2E-03 | 2.8E-01 | STD | 1.4E+00 | 1.3E+00 | 1.1E-02 | 1.3E-03 | 1.3E+00 | 1.6E-02 | 5.0E-01 | 7.3E-03 | 8.6E-02 |
| GRU | | | | | | | | | | CNN | | | | | | | | | |
| Metric | RMSE | MAE | R | MSLE | MEDAE | EVS | MAPE | MDA | RSE | Metric | RMSE | MAE | R | MSLE | MEDAE | EVS | MAPE | MDA | RSE |
| Min | 24.170 | 19.772 | 0.703 | 0.010 | 17.169 | 0.494 | 7.901 | 0.661 | 0.694 | Min | 22.414 | 18.107 | 0.629 | 0.009 | 16.003 | 0.396 | 7.439 | 0.616 | 0.597 |
| Max | 33.822 | 28.779 | 0.764 | 0.020 | 27.443 | 0.579 | 11.394 | 0.698 | 1.358 | Max | 36.234 | 30.794 | 0.676 | 0.023 | 29.537 | 0.453 | 12.134 | 0.644 | 1.559 |
| Mean | 27.705 | 22.996 | 0.745 | 0.014 | 20.921 | 0.552 | 9.140 | 0.682 | 0.927 | Mean | 29.304 | 24.201 | 0.655 | 0.015 | 21.900 | 0.427 | 9.623 | 0.626 | 1.036 |
| STD | 3.8E+00 | 3.6E+00 | 1.8E-02 | 4.1E-03 | 4.1E+00 | 2.4E-02 | 1.4E+00 | 1.1E-02 | 2.6E-01 | STD | 3.9E+00 | 3.6E+00 | 1.7E-02 | 4.1E-03 | 3.9E+00 | 2.2E-02 | 1.3E+00 | 8.3E-03 | 2.7E-01 |
| s-LSTM | | | | | | | | | | s-BiLSTM | | | | | | | | | |
| Metric | RMSE | MAE | R | MSLE | MEDAE | EVS | MAPE | MDA | RSE | Metric | RMSE | MAE | R | MSLE | MEDAE | EVS | MAPE | MDA | RSE |
| Min | 21.636 | 17.566 | 0.775 | 0.008 | 15.194 | 0.583 | 7.098 | 0.702 | 0.556 | Min | 22.690 | 18.430 | 0.773 | 0.009 | 15.796 | 0.586 | 7.393 | 0.700 | 0.611 |
| Max | 22.269 | 18.121 | 0.781 | 0.008 | 15.529 | 0.590 | 7.298 | 0.711 | 0.589 | Max | 23.532 | 19.189 | 0.777 | 0.009 | 16.725 | 0.594 | 7.674 | 0.705 | 0.658 |
| Mean | 22.004 | 17.893 | 0.777 | 0.008 | 15.393 | 0.585 | 7.218 | 0.707 | 0.575 | Mean | 23.145 | 18.833 | 0.775 | 0.009 | 16.270 | 0.590 | 7.540 | 0.702 | 0.636 |
| STD | 2.0E-01 | 1.7E-01 | 1.5E-03 | 1.4E-04 | 1.0E-01 | 2.2E-03 | 6.2E-02 | 2.4E-03 | 1.0E-02 | STD | 2.4E-01 | 2.1E-01 | 1.5E-03 | 1.8E-04 | 3.0E-01 | 2.7E-03 | 7.9E-02 | 1.6E-03 | 1.3E-02 |
| s-GRU | | | | | | | | | | l-LSTM | | | | | | | | | |
| Metric | RMSE | MAE | R | MSLE | MEDAE | EVS | MAPE | MDA | RSE | Metric | RMSE | MAE | R | MSLE | MEDAE | EVS | MAPE | MDA | RSE |
| Min | 21.441 | 17.399 | 0.770 | 0.008 | 15.173 | 0.576 | 7.039 | 0.693 | 0.546 | Min | 21.823 | 17.732 | 0.775 | 0.008 | 15.325 | 0.583 | 7.156 | 0.704 | 0.566 |
| Max | 23.411 | 19.173 | 0.782 | 0.009 | 16.666 | 0.597 | 7.687 | 0.705 | 0.651 | Max | 22.461 | 18.295 | 0.780 | 0.009 | 15.671 | 0.590 | 7.359 | 0.711 | 0.599 |
| Mean | 22.299 | 18.165 | 0.775 | 0.008 | 15.751 | 0.585 | 7.317 | 0.698 | 0.591 | Mean | 22.037 | 17.923 | 0.777 | 0.008 | 15.464 | 0.585 | 7.229 | 0.707 | 0.577 |
| STD | 5.8E-01 | 5.2E-01 | 3.4E-03 | 4.2E-04 | 5.0E-01 | 6.8E-03 | 1.9E-01 | 4.5E-03 | 3.1E-02 | STD | 1.9E-01 | 1.6E-01 | 1.5E-03 | 1.3E-04 | 9.7E-02 | 2.2E-03 | 5.8E-02 | 2.3E-03 | 9.8E-03 |
| l-BiLSTM | | | | | | | | | | l-GRU | | | | | | | | | |
| Metric | RMSE | MAE | R | MSLE | MEDAE | EVS | MAPE | MDA | RSE | Metric | RMSE | MAE | R | MSLE | MEDAE | EVS | MAPE | MDA | RSE |
| Min | 21.954 | 17.806 | 0.783 | 0.008 | 15.257 | 0.593 | 7.169 | 0.706 | 0.572 | Min | 21.133 | 17.147 | 0.776 | 0.008 | 14.796 | 0.580 | 6.957 | 0.693 | 0.530 |
| Max | 22.240 | 18.057 | 0.783 | 0.008 | 15.439 | 0.594 | 7.257 | 0.707 | 0.587 | Max | 22.185 | 18.077 | 0.779 | 0.008 | 15.630 | 0.590 | 7.284 | 0.699 | 0.585 |
| Mean | 22.062 | 17.901 | 0.783 | 0.008 | 15.336 | 0.594 | 7.202 | 0.707 | 0.578 | Mean | 21.679 | 17.646 | 0.778 | 0.008 | 15.275 | 0.585 | 7.137 | 0.696 | 0.558 |
| STD | 7.9E-02 | 7.0E-02 | 2.0E-04 | 5.6E-05 | 5.8E-02 | 4.6E-04 | 2.4E-02 | 3.2E-04 | 4.2E-03 | STD | 3.1E-01 | 2.7E-01 | 9.1E-04 | 2.1E-04 | 2.6E-01 | 2.8E-03 | 9.7E-02 | 2.0E-03 | 1.6E-02 |

**Table 5: Statistical results of ED visitor prediction for Ensemble and deep learning models**

| MLP | | | | | | | | | | DNN | | | | | | | | | |
|---|---|---|---|---|---|---|---|---|---|---|---|---|---|---|---|---|---|---|---|
| Metric | RMSE | MAE | R | MSLE | MEDAE | EVS | MAPE | MDA | RSE | Metric | RMSE | MAE | R | MSLE | MEDAE | EVS | MAPE | MDA | RSE |
| Min | 24.566 | 19.896 | 0.669 | 0.010 | 19.896 | 0.437 | 7.944 | 0.233 | 0.717 | Min | 19.691 | 15.730 | 0.758 | 0.007 | 15.730 | 0.569 | 6.486 | 0.682 | 0.461 |
| Max | 27.351 | 22.303 | 0.725 | 0.013 | 22.303 | 0.525 | 8.864 | 0.532 | 0.889 | Max | 23.281 | 18.788 | 0.770 | 0.009 | 18.788 | 0.586 | 7.537 | 0.695 | 0.644 |
| Mean | 25.353 | 20.457 | 0.695 | 0.011 | 20.457 | 0.478 | 8.141 | 0.339 | 0.764 | Mean | 20.957 | 16.854 | 0.764 | 0.008 | 16.854 | 0.578 | 6.870 | 0.689 | 0.523 |
| STD | 7.7E-01 | 7.1E-01 | 1.5E-02 | 8.2E-04 | 7.1E-01 | 2.3E-02 | 2.8E-01 | 8.8E-02 | 4.8E-02 | STD | 1.0E+00 | 8.7E-01 | 3.9E-03 | 7.1E-04 | 8.7E-01 | 5.5E-03 | 2.9E-01 | 3.9E-03 | 5.3E-02 |

| RF | | | | | | | | | | GBR | | | | | | | | | |
|---|---|---|---|---|---|---|---|---|---|---|---|---|---|---|---|---|---|---|---|
| Metric | RMSE | MAE | R | MSLE | MEDAE | EVS | MAPE | MDA | RSE | Metric | RMSE | MAE | R | MSLE | MEDAE | EVS | MAPE | MDA | RSE |
| Min | 32.342 | 26.642 | 0.327 | 0.018 | 26.642 | 0.091 | 10.624 | 0.614 | 1.243 | Min | 20.318 | 16.226 | 0.789 | 0.007 | 16.226 | 0.551 | 6.761 | 0.706 | 0.491 |
| Max | 36.273 | 29.876 | 0.492 | 0.023 | 29.876 | 0.236 | 11.863 | 0.647 | 1.563 | Max | 20.359 | 16.251 | 0.790 | 0.007 | 16.251 | 0.552 | 6.769 | 0.708 | 0.493 |
| Mean | 34.061 | 28.051 | 0.415 | 0.020 | 28.051 | 0.171 | 11.161 | 0.631 | 1.380 | Mean | 20.332 | 16.237 | 0.789 | 0.007 | 16.237 | 0.552 | 6.764 | 0.707 | 0.491 |
| STD | 1.3E+00 | 1.1E+00 | 5.4E-02 | 1.6E-03 | 1.1E+00 | 4.7E-02 | 4.0E-01 | 1.1E-02 | 1.0E-01 | STD | 1.3E-02 | 7.4E-03 | 2.4E-04 | 7.6E-06 | 7.4E-03 | 3.9E-04 | 2.5E-03 | 6.5E-04 | 6.4E-04 |

| ExT | | | | | | | | | | CatB | | | | | | | | | |
|---|---|---|---|---|---|---|---|---|---|---|---|---|---|---|---|---|---|---|---|
| Metric | RMSE | MAE | R | MSLE | MEDAE | EVS | MAPE | MDA | RSE | Metric | RMSE | MAE | R | MSLE | MEDAE | EVS | MAPE | MDA | RSE |
| Min | 23.137 | 18.857 | 0.736 | 0.009 | 18.857 | 0.456 | 7.735 | 0.708 | 0.636 | Min | 18.717 | 14.476 | 0.834 | 0.006 | 14.476 | 0.638 | 5.932 | 0.728 | 0.416 |
| Max | 23.828 | 19.434 | 0.747 | 0.010 | 19.434 | 0.474 | 7.943 | 0.728 | 0.675 | Max | 19.889 | 15.594 | 0.845 | 0.007 | 15.594 | 0.654 | 6.333 | 0.757 | 0.470 |
| Mean | 23.560 | 19.211 | 0.742 | 0.010 | 19.211 | 0.462 | 7.865 | 0.717 | 0.660 | Mean | 19.339 | 15.038 | 0.839 | 0.006 | 15.038 | 0.644 | 6.129 | 0.741 | 0.445 |
| STD | 2.1E-01 | 1.9E-01 | 3.2E-03 | 1.6E-04 | 1.9E-01 | 5.5E-03 | 6.9E-02 | 6.0E-03 | 1.2E-02 | STD | 3.6E-01 | 3.5E-01 | 3.5E-03 | 2.1E-04 | 3.5E-01 | 5.6E-03 | 1.3E-01 | 8.3E-03 | 1.7E-02 |

| AdaB | | | | | | | | | | XGB | | | | | | | | | |
|---|---|---|---|---|---|---|---|---|---|---|---|---|---|---|---|---|---|---|---|
| Metric | RMSE | MAE | R | MSLE | MEDAE | EVS | MAPE | MDA | RSE | Metric | RMSE | MAE | R | MSLE | MEDAE | EVS | MAPE | MDA | RSE |
| Min | 22.914 | 18.637 | 0.672 | 0.009 | 18.637 | 0.396 | 7.684 | 0.651 | 0.624 | Min | 29.735 | 24.452 | 0.473 | 0.015 | 24.452 | 0.140 | 9.896 | 0.375 | 1.051 |
| Max | 25.034 | 20.530 | 0.720 | 0.011 | 20.530 | 0.472 | 8.431 | 0.686 | 0.745 | Max | 31.312 | 25.865 | 0.592 | 0.017 | 25.865 | 0.244 | 10.382 | 0.488 | 1.165 |
| Mean | 23.827 | 19.395 | 0.698 | 0.010 | 19.395 | 0.434 | 7.994 | 0.670 | 0.675 | Mean | 30.435 | 25.068 | 0.541 | 0.016 | 25.068 | 0.187 | 10.120 | 0.435 | 1.101 |
| STD | 6.3E-01 | 5.8E-01 | 1.5E-02 | 5.0E-04 | 5.8E-01 | 2.2E-02 | 2.2E-01 | 1.1E-02 | 3.6E-02 | STD | 5.8E-01 | 5.5E-01 | 3.5E-02 | 5.4E-04 | 5.5E-01 | 3.9E-02 | 1.7E-01 | 3.0E-02 | 4.2E-02 |

| LGB | | | | | | | | | | HGBR | | | | | | | | | |
|---|---|---|---|---|---|---|---|---|---|---|---|---|---|---|---|---|---|---|---|
| Metric | RMSE | MAE | R | MSLE | MEDAE | EVS | MAPE | MDA | RSE | Metric | RMSE | MAE | R | MSLE | MEDAE | EVS | MAPE | MDA | RSE |
| Min | 21.324 | 16.988 | 0.801 | 0.008 | 16.988 | 0.574 | 6.916 | 0.725 | 0.540 | Min | 20.309 | 16.220 | 0.789 | 0.007 | 16.220 | 0.551 | 6.758 | 0.706 | 0.490 |
| Max | 21.418 | 17.080 | 0.806 | 0.008 | 17.080 | 0.585 | 6.969 | 0.734 | 0.545 | Max | 20.359 | 16.251 | 0.790 | 0.007 | 16.251 | 0.553 | 6.769 | 0.708 | 0.493 |
| Mean | 21.368 | 17.033 | 0.804 | 0.008 | 17.033 | 0.580 | 6.942 | 0.730 | 0.543 | Mean | 20.334 | 16.238 | 0.789 | 0.007 | 16.238 | 0.552 | 6.764 | 0.707 | 0.491 |
| STD | 3.7E-02 | 3.6E-02 | 2.0E-03 | 3.5E-05 | 3.6E-02 | 3.9E-03 | 1.8E-02 | 2.6E-03 | 1.9E-03 | STD | 1.4E-02 | 1.0E-02 | 3.2E-04 | 8.7E-06 | 1.0E-02 | 4.9E-04 | 3.7E-03 | 7.0E-04 | 7.0E-04 |

Table 6 presents a detailed analysis of the performance metrics associated with two distinct neural network models, namely the MLP and the DNN, alongside two tree-based models, which include Random Forest (RF) and Extra Trees, as well as an impressive collection of six ensemble models, all evaluated across a total of nine different assessment metrics that provide a comprehensive view of their efficacy. Upon careful examination of the prediction results, it becomes abundantly clear that the standout winner among these models was none other than CatBoost, which astonishingly achieved an accuracy rate of 0.839 and a Mean Absolute Error (MAE) of 15.04, showcasing its remarkable capabilities in the realm of predictive modelling. This exceptional performance can likely be attributed to CatBoost's innovative ordered boosting methodology, specifically designed to minimise overfitting by ensuring that any potential leakage of target information is drastically reduced, thereby enhancing the overall integrity of the model. The ordered boosting technique employed by CatBoost adeptly addresses this significant issue, resulting in a model that not only produces more accurate predictions but also demonstrates a higher level of reliability, particularly in contexts where either the size or quality of the data poses challenges to effective analysis. Furthermore, it is worth noting that the DNN outperformed their MLP counterparts significantly, a distinction that can be attributed to their superior network depth, which enables them to effectively capture intricate relationships within the data, along with their advanced capabilities in feature extraction and a more robust application of regularisation techniques that help prevent overfitting and enhance model performance.

Also, we tested the effectiveness of three different master-learners: MLP, linear regression, and Ridge regression on the proposed Meta-ED prediction model. The relevant results are in Table 6. We can see that a combination of MLP with four sub-learners (Meta-ED-MLP) outperformed the other two models by 2% on average in terms of accuracy.

Table 6: Statistical results of ED visitor prediction for three types of the proposed Meta-ED models

| Meta-ED-MLP | | | | | | | | | |
|---|---|---|---|---|---|---|---|---|---|
| Metric | RMSE | MAE | R | MSLE | MEDAE | EVS | MAPE | MDA | RSE |
| Min | 15.969 | 12.016 | 0.850 | 0.005 | 12.016 | 0.684 | 5.129 | 0.738 | 0.303 |
| Max | 19.598 | 15.886 | 0.866 | 0.007 | 15.886 | 0.738 | 6.943 | 0.757 | 0.456 |
| Mean | 17.366 | 13.566 | 0.856 | 0.006 | 13.566 | 0.714 | 5.915 | 0.748 | 0.360 |
| STD | 1.1E+00 | 1.1E+00 | 4.2E-03 | 6.9E-04 | 1.1E+00 | 1.6E-02 | 5.1E-01 | 5.5E-03 | 4.6E-02 |
| Meta-ED-LR | | | | | | | | | |
| Metric | RMSE | MAE | R | MSLE | MEDAE | EVS | MAPE | MDA | RSE |
| Min | 21.853 | 17.709 | 0.831 | 0.009 | 17.709 | 0.677 | 7.766 | 0.725 | 0.567 |
| Max | 24.911 | 21.135 | 0.847 | 0.011 | 21.135 | 0.713 | 9.206 | 0.754 | 0.737 |
| Mean | 23.545 | 19.581 | 0.836 | 0.010 | 19.581 | 0.691 | 8.564 | 0.746 | 0.659 |
| STD | 8.1E-01 | 9.0E-01 | 4.9E-03 | 5.9E-04 | 9.0E-01 | 9.8E-03 | 3.8E-01 | 6.7E-03 | 4.5E-02 |
| Meta-ED-Ridge | | | | | | | | | |
| Metric | RMSE | MAE | R | MSLE | MEDAE | EVS | MAPE | MDA | RSE |
| Min | 22.655 | 18.755 | 0.825 | 0.009 | 18.755 | 0.669 | 8.184 | 0.731 | 0.610 |
| Max | 24.239 | 20.439 | 0.847 | 0.011 | 20.439 | 0.707 | 8.938 | 0.757 | 0.698 |
| Mean | 23.680 | 19.708 | 0.835 | 0.010 | 19.708 | 0.687 | 8.632 | 0.746 | 0.667 |
| STD | 4.8E-01 | 5.0E-01 | 6.7E-03 | 4.1E-04 | 5.0E-01 | 1.2E-02 | 2.2E-01 | 8.0E-03 | 2.7E-02 |

The analysis conducted on 23 deep learning and machine learning models to predict the daily number of ED visitors is detailed in Figure 9. Each model was trained and tested in ten independent iterations, culminating in a comprehensive evaluation of their performance showcased within the depicted boxplot. In terms of prediction accuracy, as indicated by the R-value (refer to Figure 9 (a)), the standout performer was the CatBoost (CatB) Ensemble model, exhibiting an impressive average accuracy of 84%. Following closely behind, LightBoost (LGB) and Gradient Boosting (GBR) secured the second and third positions with 80% and 79% accuracy rates, respectively. In evaluating predictive models using the Explained Variance Score (EVS) as the second metric (refer to Figure 9 (b)), which quantifies the predictability of the dependent variable from the independent variables, significant insights emerged. The EVS metric spans a range from negative infinity to 1, where a score of 1 signifies excellent prediction accuracy, while lower scores denote decreasing predictive precision. Remarkably, CatBoost emerged as the top performer, showcasing an impressive average EVS of 0.64, surpassing other models by a notable margin. Specifically, CatBoost's EVS outperformed LightBoost (LGB) by 10.3%, underscoring its superior ability to capture and explain the variance in the dataset, thereby enhancing its predictive power significantly in this analytical context. In the assessment of Mean Absolute Error (MAE) and Mean Squared Logarithmic Error (MSLE), where lower values signify superior performance and reduced training errors, noteworthy observations were made. Illustrated in Figure 9 (c) and (d), CatBoost particularly outshined all other 22 machine learning models in these metrics, reaffirming its exceptional performance. This unique advantage of CatBoost, which can be attributed to its utilisation of an advanced variant of gradient boosting that incorporates a more potent form of regularisation, is truly enlightening. This regularisation technique mitigates overfitting, enhancing the model's generalisation capabilities. Such expertise is particularly advantageous in scenarios with limited data, such as our dataset comprising 8600 samples. CatBoost's ability to prevent overfitting and bolster generalisation performance sets it apart, leading to superior predictive accuracy and model robustness.

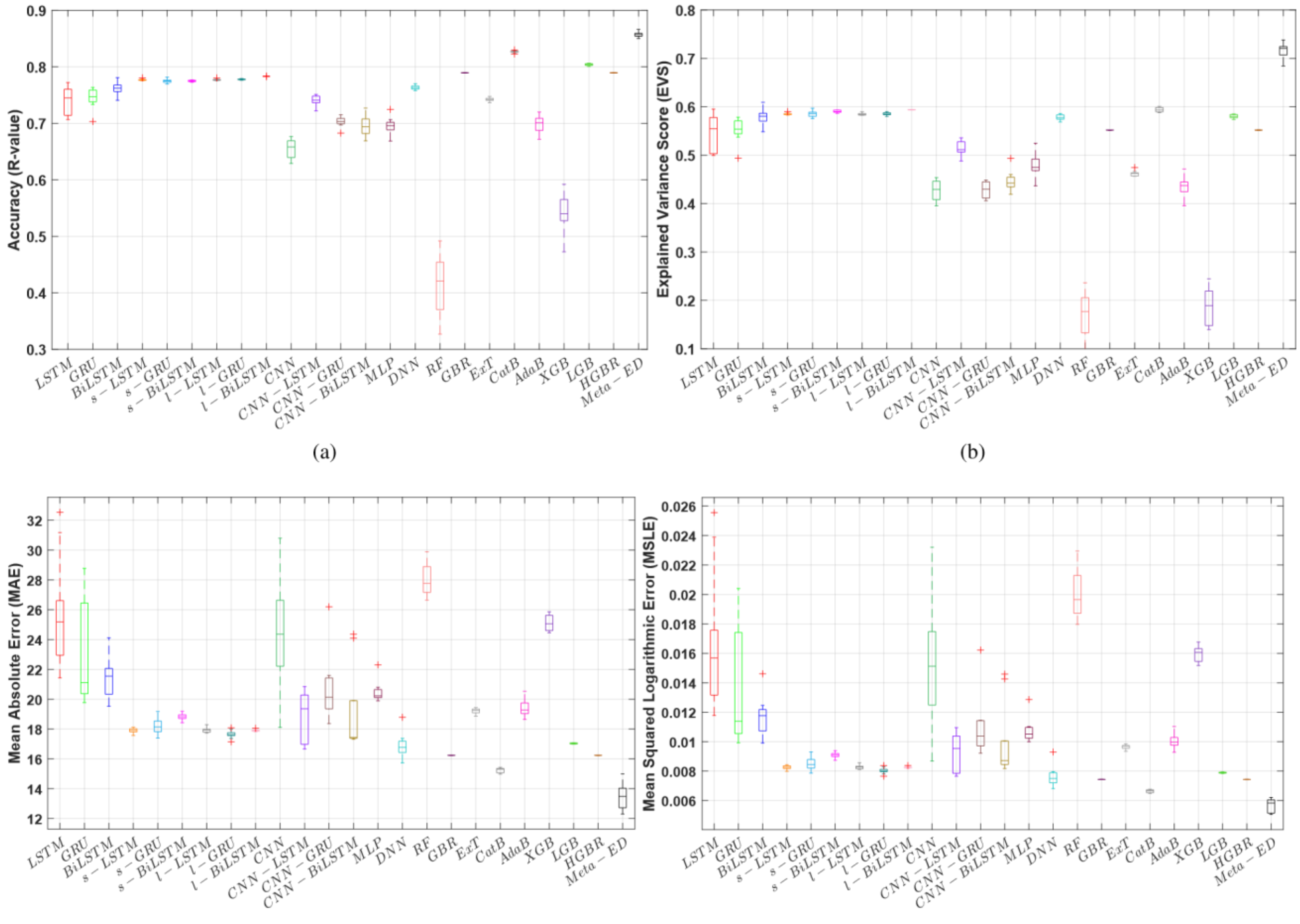

(a) (b)

(c) (d)

**Figure 9: The statistical results for the proposed hybrid model's performance compared with 23 ML, DL and Ensemble models in predicting ED visitors number (daily) based on collected data from Canberra Hospital.**

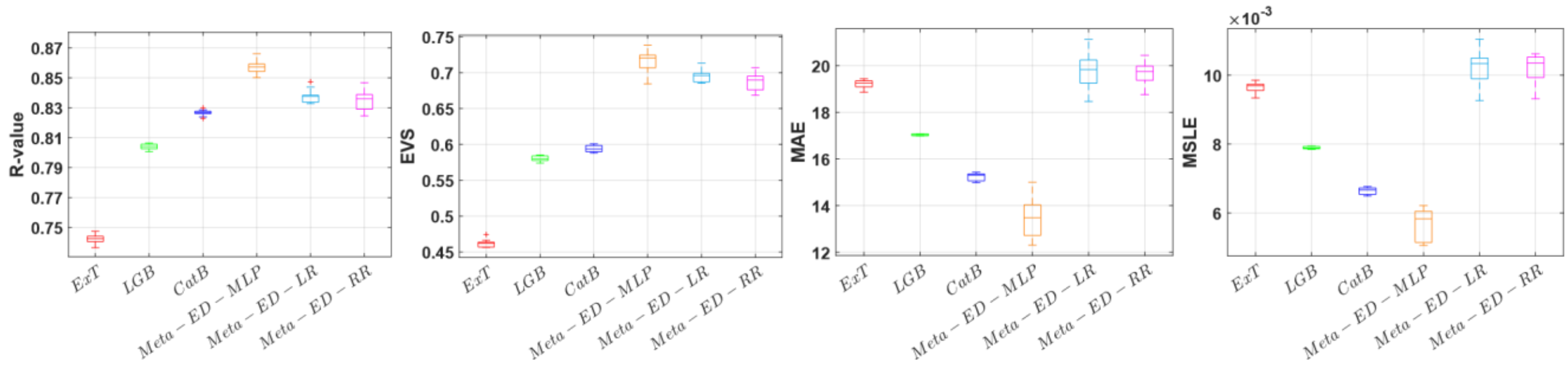

(a) (b) (c) (d)

**Figure 10: The statistical prediction results for the proposed Meta-learning models compared with their components (ExT, LGB, and CatB) in predicting ED visitor number (daily) based on collected data from Canberra Hospital.**

To evaluate the benefits of the proposed Meta-ED model, incorporating various master learners (MLP, LR, RR), against individual ensemble models such as CatBoost, LightGBM, and Extra Trees, we conducted a comparative experiment by running each method ten times. The results, displayed in Figure 10, demonstrate that Meta-ED-MLP outperformed other Meta models as well as ensemble models in terms of R-value, Explained Variance Score (EVS), Mean Absolute Error (MAE), and Mean Squared Logarithmic Error (MSLE) on average.

To facilitate a comparative visualisation of the proposed Meta-ED learner against other widely used ensemble models, Figure 11 is presented. The results clearly demonstrate that the Meta-ED model excels in capturing the complex and dynamic patterns of ED visitor numbers, as evidenced by testing data from 2019 to 2022. The zoomedin segments of the overall line chart further highlight the Meta-ED model's superior accuracy in predicting various fluctuation patterns in visitor numbers. Besides, both CatBoost and LGBoost exhibit comparable performance, effectively predicting the upward and downward trends in ED crowding. However, they do not match the precision of the Meta-ED model.

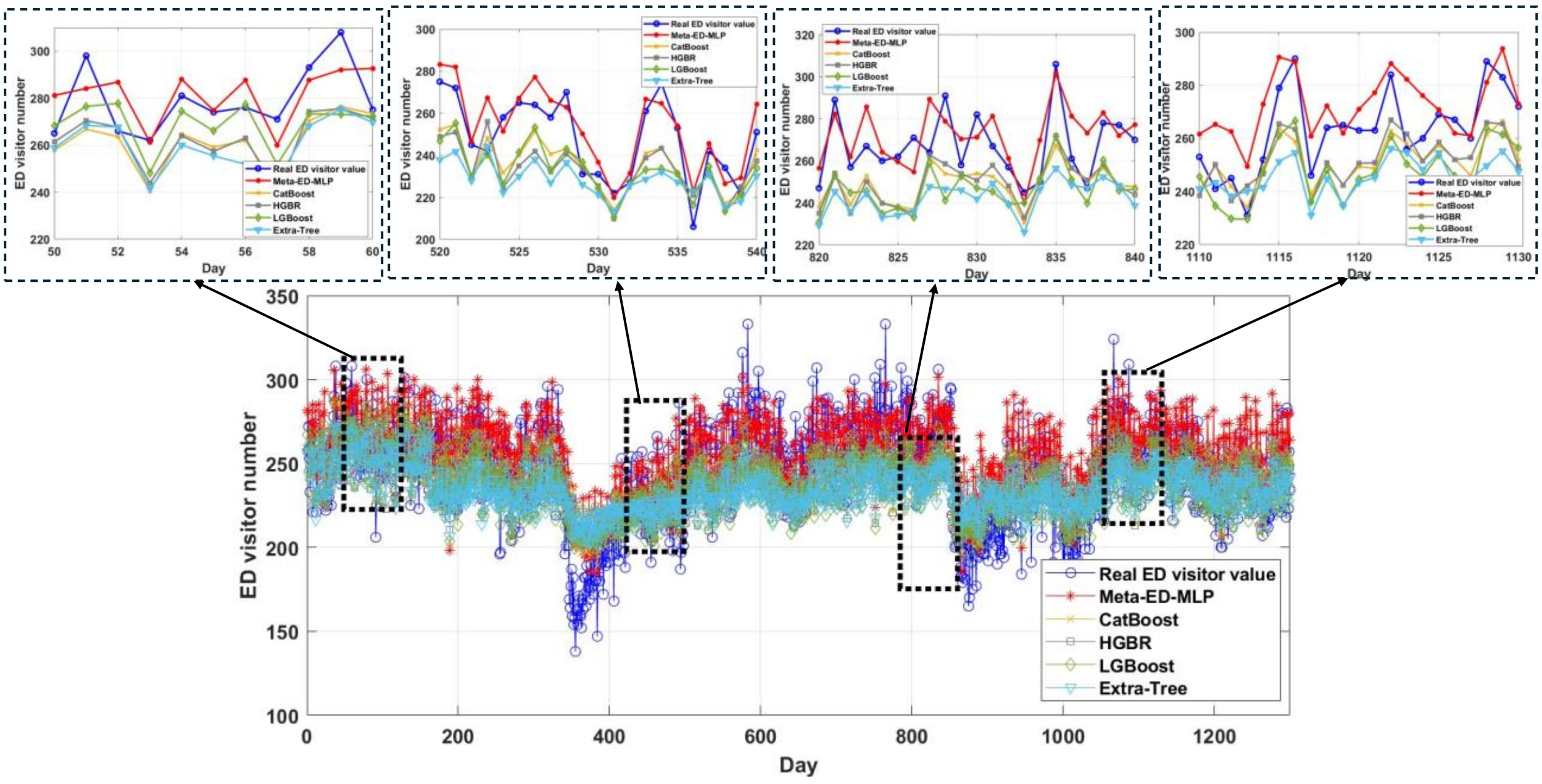

**Figure 11: Performance comparison between the proposed Meta-ED and other well-known Ensemble models with the real ED visitor number.**

The results presented in Table 7 indicate a comprehensive comparison of the proposed Meta-ED model against 23 other machine learning models evaluated using the Friedman test. Notably, the Meta-ED model achieved the highest average rank in performance (R-value), underscoring its superior predictive capabilities relative to the competing models. The p-value results further substantiate this finding, demonstrating that the performance of the Meta-ED model is statistically significantly different from that of the other models. This suggests that the improvements offered by the Meta-ED model are not only substantial but also reliable, highlighting its potential as a robust solution for the targeted prediction tasks. The combination of top-ranking performance and significant p-values positions the Meta-ED model as a leading choice in the evaluated set of machine learning models.

**Table 7: Comparative Analysis of 23 Machine Learning Models and the Proposed Meta-ED Model Using Friedman Test: P-values and Average** Ranks

| Model | LSTM | GRU | BiLSTM | s-LSTM | s-GRU | s-BiLSTM | l-LSTM | l-GRU |
|---|---|---|---|---|---|---|---|---|
| Avg-Rank | 14.6 | 15.4 | 12.9 | 9 | 9.9 | 10.1 | 8.7 | 7.9 |
| P-value | 2.03E-13 | 3.05E-15 | 9.01E-18 | 3.55E-24 | 8.07E-23 | 2.12E-24 | 3.18E-24 | 2.09E-24 |
| Model | l-BiLSTM | CNN | CNN-LSTM | CNN-GRU | CNN-BiLSTM | MLP | DNN | RF |
| Avg-Rank | 6 | 21.8 | 15.7 | 18.8 | 19.5 | 19.9 | 12.5 | 24 |
| P-value | 5.62E-24 | 8.54E-21 | 5.04E-21 | 6.28E-23 | 1.09E-17 | 1.36E-19 | 1.78E-23 | 1.02E-17 |
| Model | GBR | ExT | CatB | AdaB | XGB | LGB | HGBR | Meta-ED-MLP |
| Avg-Rank | 4.5 | 15.6 | 2 | 19.7 | 23 | 3 | 4.5 | 1 |
| P-value | 3.14E-23 | 6.57E-26 | 1.44E-15 | 1.34E-19 | 5.09E-19 | 2.32E-20 | 3.17E-23 | N/A |

### *3.6. Feature Selection Results*

We employed the widely used and efficient Recursive Feature Elimination (RFE) method to identify the optimal subset of practical features for prediction. Two key factors were considered during the feature selection phase: the number of features and the specific subset of features chosen. We conducted a comprehensive evaluation of the performance of RFE with varying feature counts, ranging from 15 to 69 (all available features). Figure 12 illustrates the accuracy of these feature selection experiments and the mean absolute error (MAE). The results indicate that the optimal number of features, in terms of accuracy, is 61. However, the lowest error was observed with 30 and 40 features when focusing on validation MAE.

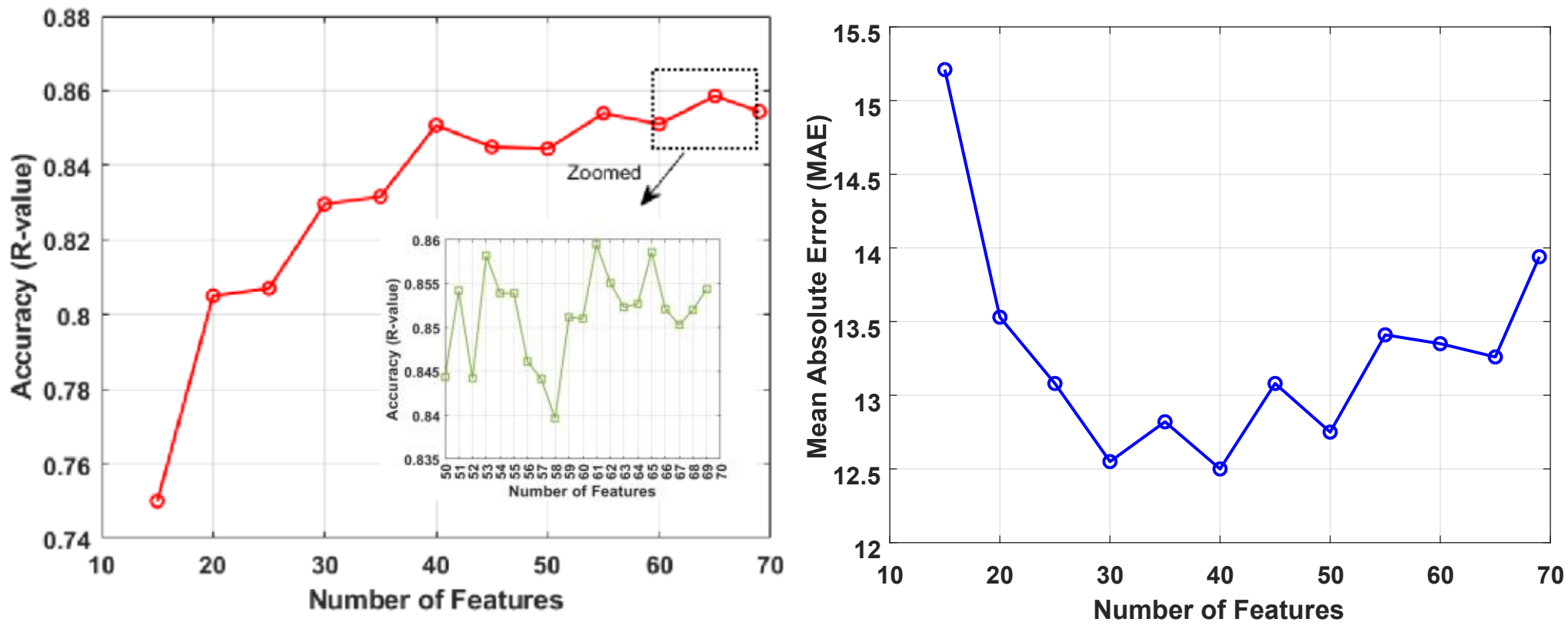

(a) (b)

**Figure 12: The performance of Recursive Feature Elimination (RFE) in optimising the subset of features based on the Meta-ED model.**

Table 8 shows that the Meta-ED analyses encompass a series of distinct perspectives on ED visitor prediction. Meta-ED-1 delves into forecasting without temporal markers like year, month, and day. Conversely, Meta-ED-2 explores predictions void of disposition parameters, condensing the intricate variability into a more streamlined outcome. In a similar vein, Meta-ED-3 studies performance without age group influences, offering a purer insight into prediction dynamics. Moving along, Meta-ED-4 isolates the impact of 12 prevalent ICD10 diagnoses from the equation, refining the predictive landscape. Furthermore, Meta-ED-5 extends this observation by excluding the top 5 daily ICD10 occurrences, challenging conventional forecasting norms. Transitioning further, Meta-ED-6 reveals outcomes without the weight of ten frequently recurring ICD10 codes, reshaping predictive accuracy. Strikingly, Meta-ED-7 strips away the reliance on triage for predictions, unveiling a novel dimension in forecasting methodologies. Lastly, Meta-ED-8 embraces the entirety of 69 features, offering a comprehensive lens into the multifaceted realm of ED visitor prognosis. After evaluating the performance of Meta-ED by removing specific features (from Meta-ED-1 to Meta-ED-7 as illustrated in Figure 13), the reduction in prediction accuracy compared to Meta-ED-8 (which utilises the full set of features) is noteworthy, with declines of 87%, 1.0%, 1.66%, 5.76%, 39.25%, 3.25%, and 2.66%, respectively. The results indicate that the most critical feature is temporal markers, including year, month, and day (Meta-ED-1). The absence of these temporal features severely impacts the model's ability to forecast ED visitors, as it removes essential seasonal trends, such as those associated with flu outbreaks in winter. Additionally, neglecting temporal context can obscure evolving visitor patterns influenced by factors like population growth and changes in healthcare access, thereby eliminating valuable insights necessary for accurate predictions.

**Table 8: Effect of Feature Selection on Meta-ED Model Performance in Predicting Daily ED Visitor Numbers. The "Impact" column quantifies the reduction in prediction accuracy observed after the removal of specific features from the Meta-ED model.**

| Model | Details | Accuracy | Impact (%) |
|---|---|---|---|
| Meta-ED-1 | without temporal features (Year, Month, and Day) | 45.9% | 86.97 |
| Meta-ED-2 | without disposition | 84.9% | 0.96 |
| Meta-ED-3 | without age groups | 84.3% | 1.66 |
| Meta-ED-4 | without most viral ICD10 features | 81.1% | 5.76 |
| Meta-ED-5 | without most frequent five ICD10 features (daily) | 61.6% | 39.25 |
| Meta-ED-6 | without most frequent ICD10 features | 76.7% | 11.82 |
| Meta-ED-7 | without triage features | 83.5% | 2.66 |
| Meta-ED-8 | without climate parameters | 83.0% | 3.25 |
| Meta-ED-9 | All features | 85.7% | – |

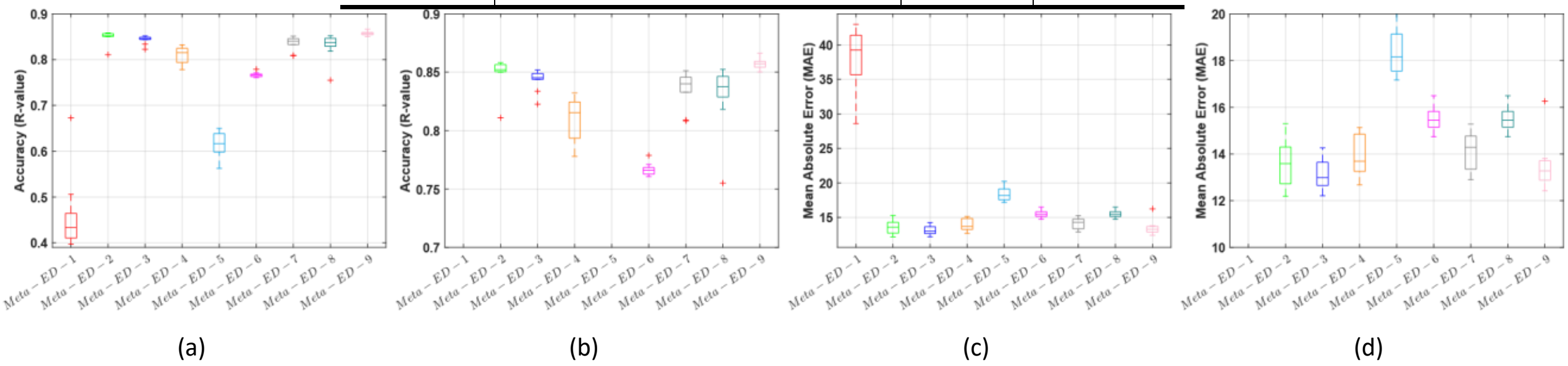

**Figure 13: Impact of Feature Selection on Proposed Meta-ED Model Performance. (a) Accuracy, (b) Zoomed Version of Accuracy, (c) MAE (Mean Absolute Error), (d) Zoomed Version of MAE.**

### *3.7. Hyper-parameters tuning*

The fine-tuning of hyper-parameters is essential for effectively training contemporary ML models and has a significant impact on boosting their efficacy. It also aids in navigating the complexities and frameworks of these algorithms [69]. Thus, choosing hyper-parameters before fitting models to a dataset judiciously is not just important; it's imperative. Techniques that adjust based on data, like cross-validation, are frequently employed to refine and assess models using separate training and testing datasets. Classic hyper-parameter search techniques encompass grid search and random search. Furthermore, recent studies have unveiled more sophisticated methodologies, including Bayesian optimisation and meta-heuristic algorithms, which are specifically designed to tackle unique challenges for enhanced efficiency and precision.

To explore the hyper-parameter search space, we selected the two best-performing models: CatBoost and LightGBM. Using grid search for CatBoost, we evaluated four key hyper-parameters: maximum tree depth, learning rate, sub-sample rate, and L2 regularisation. The hyper-parameters for LightGBM were similar, with the exception of L2 regularisation, which was replaced by the number of estimators. These hyper-parameters were chosen to optimise the models' predictive performance by fine-tuning their complexity and learning capabilities. Figure 14 illustrates the performance landscape of CatBoost and LightGBM based on accuracy and MAE. As shown in Figure 14 (a) and (e), the learning rate and tree depth have a more significant impact compared to other hyper-parameters, leading to an improvement of up to 20% in terms of accuracy.

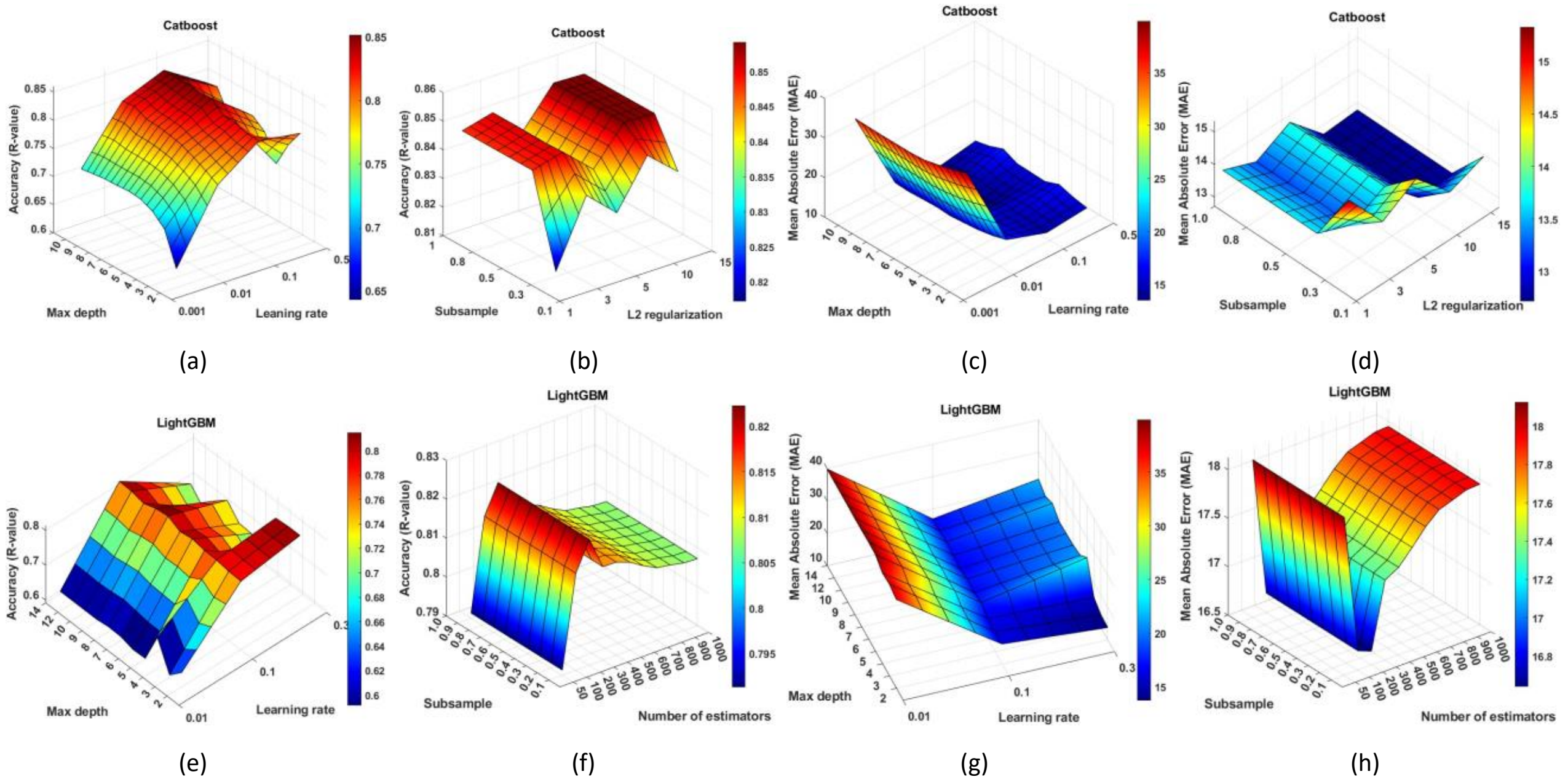

**Figure 14: Visualising performance landscape: Grid Search optimisation of hyper-parameters for CatBoost and LightGBM models based on Accuracy (R-value) and Mean Absolute Error (MAE).**

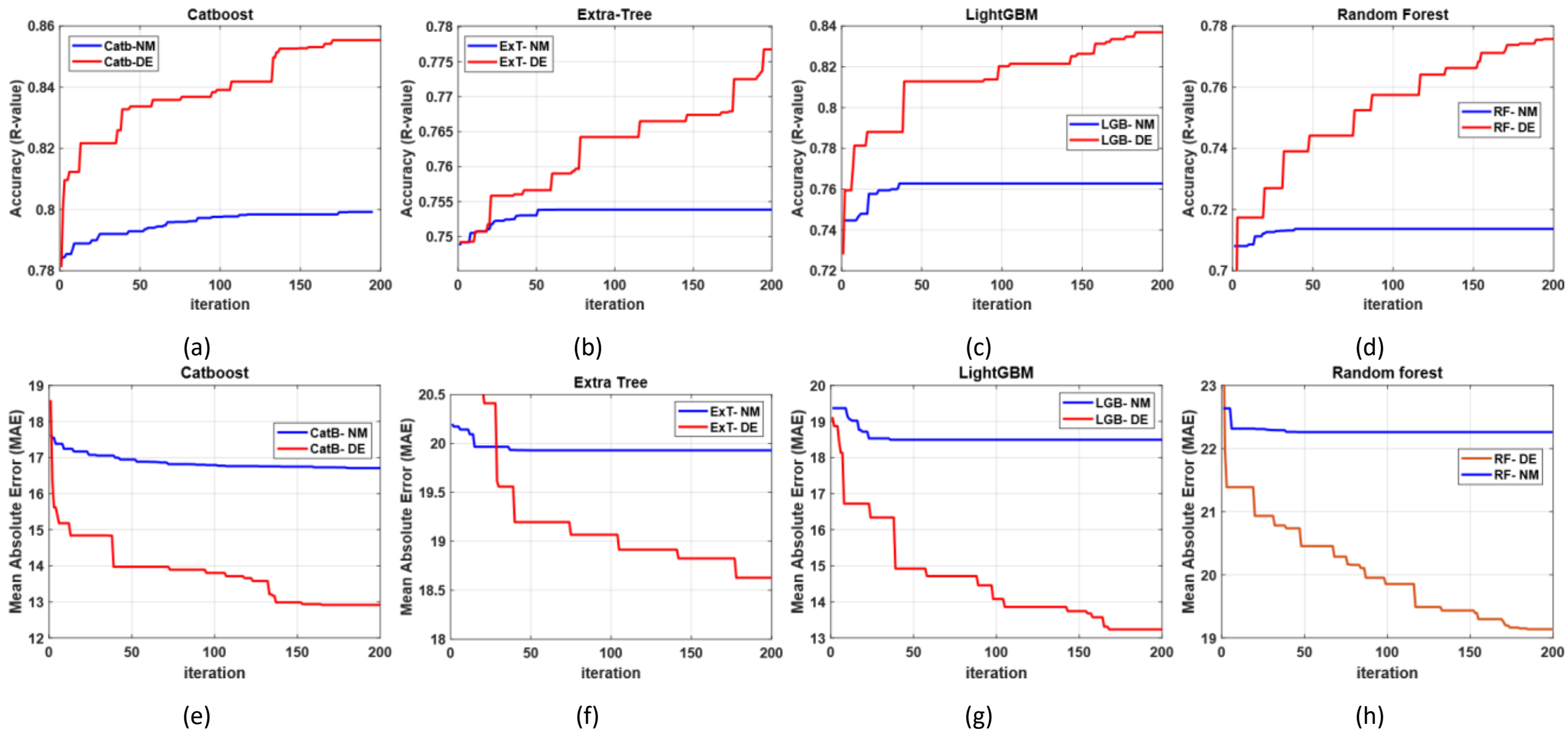

**Figure 15: Performance comparison of Nelder-Mead and Differential Evolution in Hyper-parameter Optimisation across a, e) CatBoost, b, f) Extra Trees, c, g) LightGBM, and d,h) Random Forest**

Figure 15 presents a performance comparison of two optimization methods, Nelder-Mead and Differential Evolution (DE), for tuning the hyper-parameters of the Meta-ED model. The most notable observation is the superior performance of DE in finding optimal settings compared to Nelder-Mead. Nelder-Mead faced premature convergence and became stuck in a local optimum after only a few iterations, lacking an effective strategy to escape these sub-optimal regions.

**Table 9: Best-found hyper-parameters configuration of four predictive models.**

| LightGBM | | | | | | |
|---|---|---|---|---|---|---|
| Hyper | Estimation number | Max depth | Learning rate | Sub-sample | Leave number | Accuracy |
| 1 | 615 | 10 | 0.07 | 0.66 | 21 | 0.8442 |
| 2 | 469 | 2 | 0.33 | 0.52 | 54 | 0.8400 |
| 3 | 1587 | 3 | 0.04 | 0.13 | 110 | 0.8396 |
| 4 | 661 | 2 | 0.28 | 0.67 | 30 | 0.8399 |
| 5 | 1712 | 3 | 0.03 | 0.55 | 37 | 0.8399 |
| Catboost | | | | | | |
| Hyper | Estimation number | Max depth | Learning rate | Sub-sample | L2 | Accuracy |
| 1 | 395 | 6 | 0.30 | 0.87 | 57 | 0.8560 |
| 2 | 488 | 7 | 0.22 | 0.55 | 24 | 0.8582 |
| 3 | 488 | 7 | 0.15 | 0.43 | 11 | 0.8578 |
| 4 | 467 | 6 | 0.20 | 0.53 | 42 | 0.8572 |
| 5 | 421 | 6 | 0.26 | 0.20 | 38 | 0.8563 |
| Random forest | | | | | | |
| Hyper | Estimation number | Max depth | min samples split | min samples leaf | max features | Accuracy |
| 1 | 110 | 7 | 9 | 3 | 0.447 | 0.7803 |
| 2 | 223 | 12 | 4 | 2 | 0.352 | 0.7799 |
| 3 | 402 | 13 | 5 | 1 | 0.413 | 0.7763 |
| 4 | 320 | 14 | 10 | 3 | 0.286 | 0.7759 |
| 5 | 259 | 12 | 11 | 4 | 0.312 | 0.7744 |
| Extra Tree | | | | | | |
| Hyper | Estimation number | Max depth | min samples split | min samples leaf | max features | Accuracy |
| 1 | 145 | 12 | 7 | 2 | 0.294 | 0.7800 |
| 2 | 358 | 14 | 2 | 1 | 0.33 | 0.7797 |
| 3 | 236 | 12 | 8 | 3 | 0.455 | 0.7781 |
| 4 | 428 | 13 | 12 | 2 | 0.355 | 0.7773 |
| 5 | 425 | 14 | 8 | 2 | 0.37 | 0.7763 |

### *3.8. Explainable AI and feature importance*

As illustrated in Figure 16, the most influential feature is the year of attendance at the ED, which has a significant positive impact on predicting visitor numbers. Following this, the features ICD10 Z53.1 (did not wait for treatment) and the five most frequent daily diagnoses (first ICD-10, second ICD-10, etc.) exhibit an impact range of -10 to +10, with lower values indicating a negative contribution to the predictions. Other features have a more modest influence, with impacts ranging between -5 and +5.

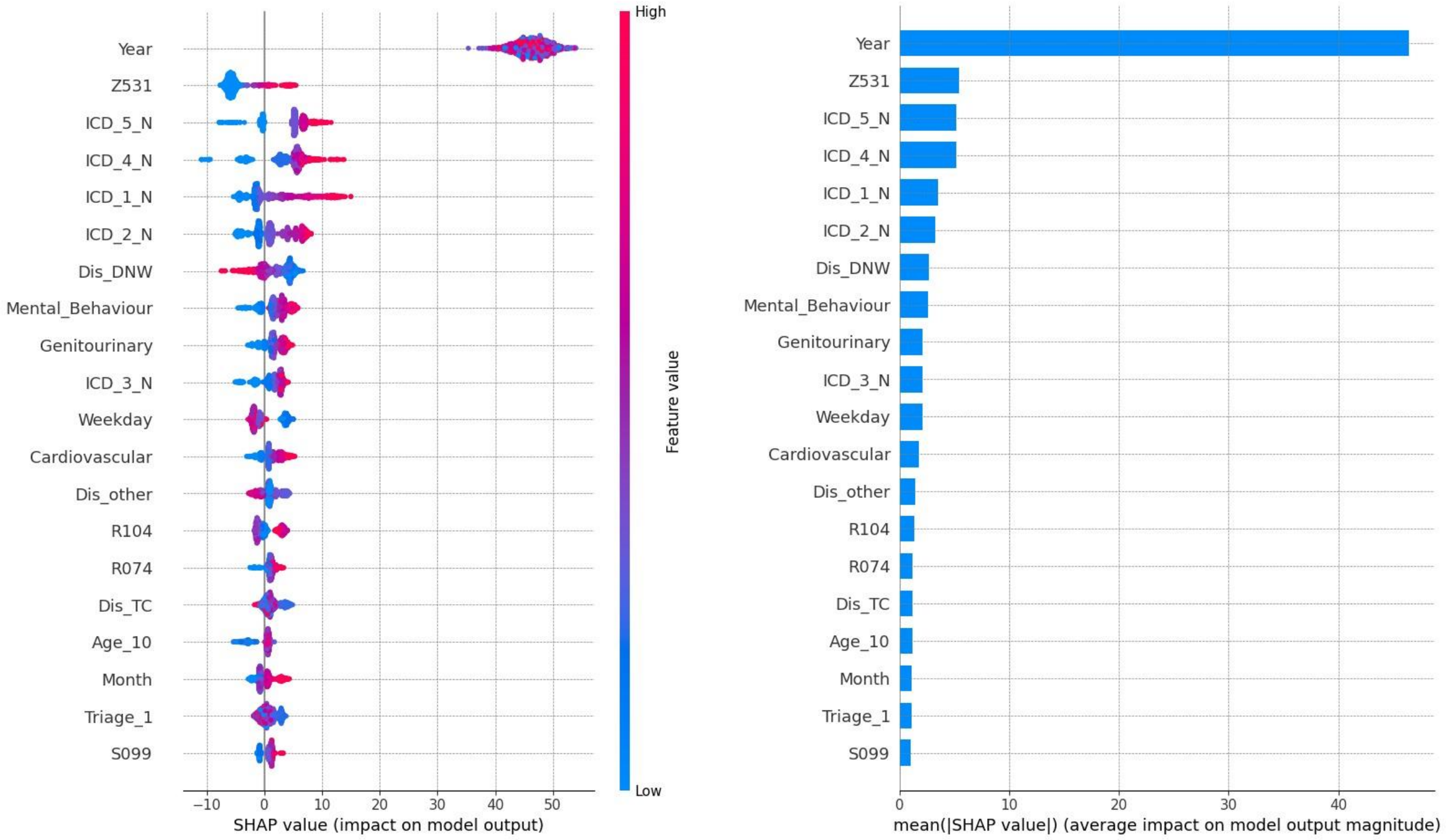

(a) (b)

**Figure 16: Summary of SHAP values illustrating the importance of various features in predicting the daily number of ED visitors. (a) The distribution of SHAP values for each feature highlights their respective contributions to model predictions, revealing how each feature influences the expected volume of ED visitors. (b) The accompanying bar plot displays the mean absolute SHAP value for each feature across all instances in the test dataset. Higher mean absolute values indicate greater contributions to predicting ED visitor numbers, allowing for a clear comparison of feature significance in driving patient inflow.**

Figure 17 illustrates the contributions of various features, which can be either positive or negative, in predicting the number of Emergency Department (ED) visitors based on four random samples from the test dataset. As anticipated, the Year feature shows the highest positive contribution, ranging from 45 to 50. Additionally, the fourth and fifth most frequent ICD-10 codes rank as the second most influential features, contributing between five and ten across different samples. The blue-coloured features indicate a negative contribution to the final prediction values, such as ICD-10 code Z53.1.

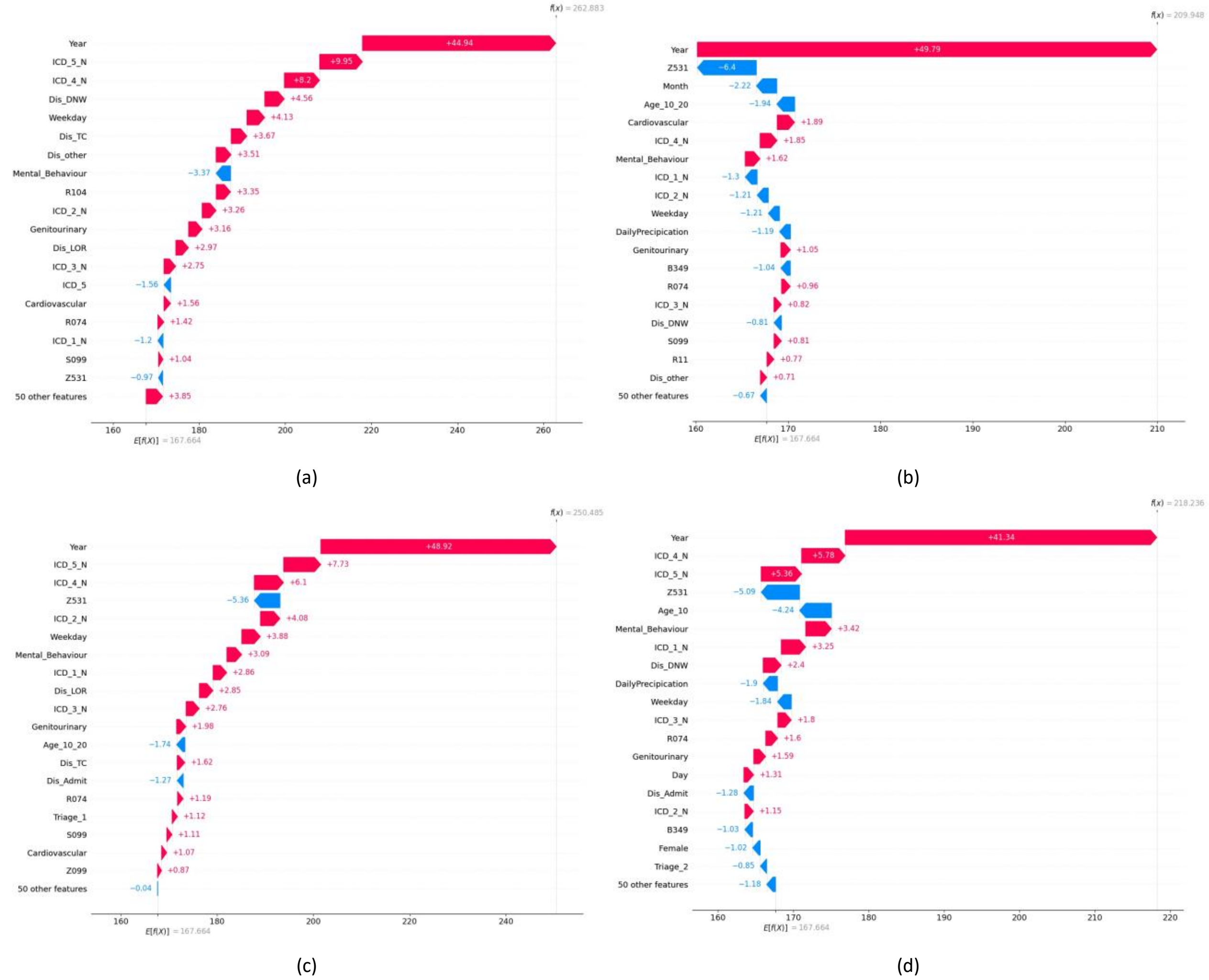

**Figure 17: The waterfall plot illustrates the contributions of individual features to the model's predictions for ED visitor numbers across four random samples from the test dataset. This visualisation aids in understanding how different features influence the predictions, either increasing or decreasing them relative to a baseline value. Z53.1 (Did not wait for treatment), R10.4 (Pain in abdomen),**

We conducted an analysis to examine how the values of climate features interact to influence the model's output. The results of this analysis are presented in Figure 18. The x-axis represents the values of the analyzed feature, while the y-axis displays the corresponding SHAP values. A positive SHAP value indicates that the feature contributes positively to the prediction, whereas a negative SHAP value signifies a negative contribution. For instance, as illustrated in Figure 18(a), low values of maximum daily temperature are associated with negative SHAP values, suggesting that colder weather may result in fewer ED visits. This phenomenon could be explained by reduced outdoor activity levels or decreased heat-related health issues on colder days. Conversely, higher maximum daily temperatures correspond to positive SHAP values, indicating that the predicted number of ED visitors also increases as temperatures rise. This trend may be linked to a rise in heat-related health issues, such as heat exhaustion, dehydration, and exacerbations of chronic conditions, which tend to be more prevalent during warmer weather.

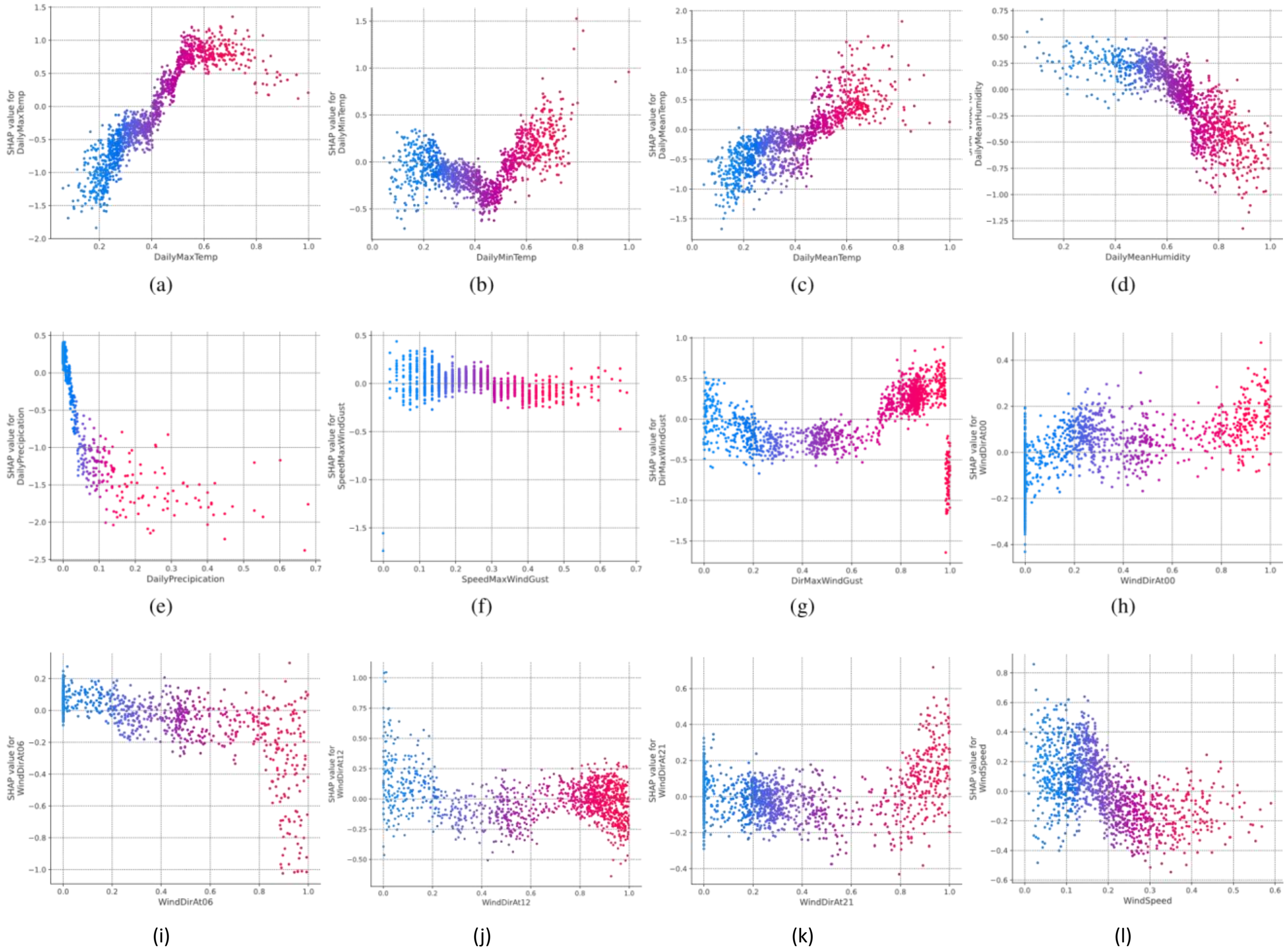

**Figure 18: Evaluating the importance of climate parameters in predicting the daily number of ED visitors using SHAP (Shapley Additive exPlanations) values.**

The visualisation in Figure 19 shows the varying contribution levels of the primary reasons for ED visits. With the exception of the N39.0 ICD-10 code representing Urinary Tract Infections, other frequent diagnosis codes within the ICD-10 system positively impact ED visitor numbers by influencing patient volumes. Conversely, a decrease in these rates results in a negative impact. An analysis reveals the differential effects of ICD-10 codes on ED visitor rates ranging from -8 to +6. Building upon the preceding analysis, Z53.1 exhibits the most significant positive (+6) and negative (-8) contributions to the output model. Cardiovascular and Genitourinary parameters are closely followed, displaying positive and negative impact values at +5 and -3.

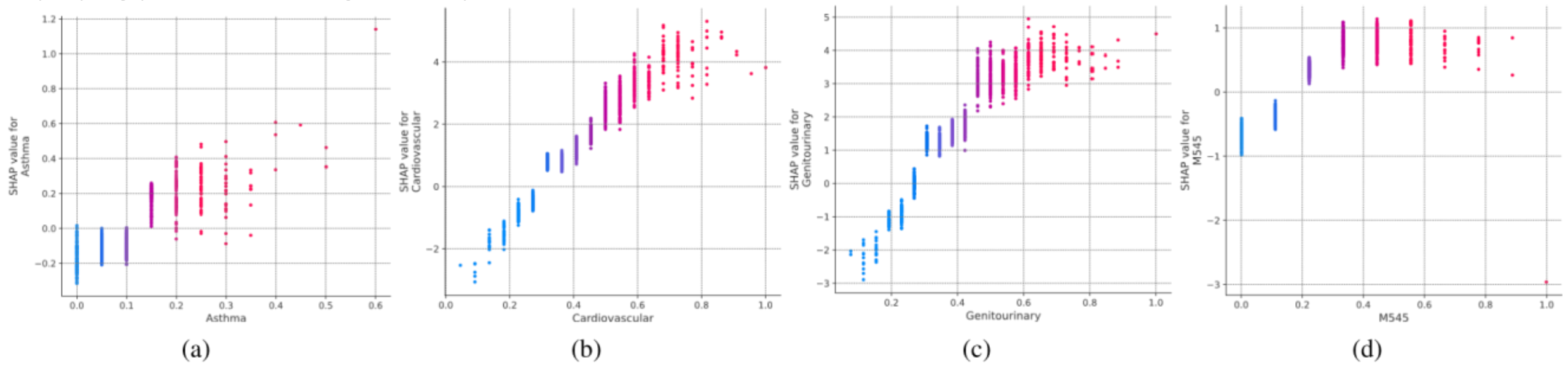

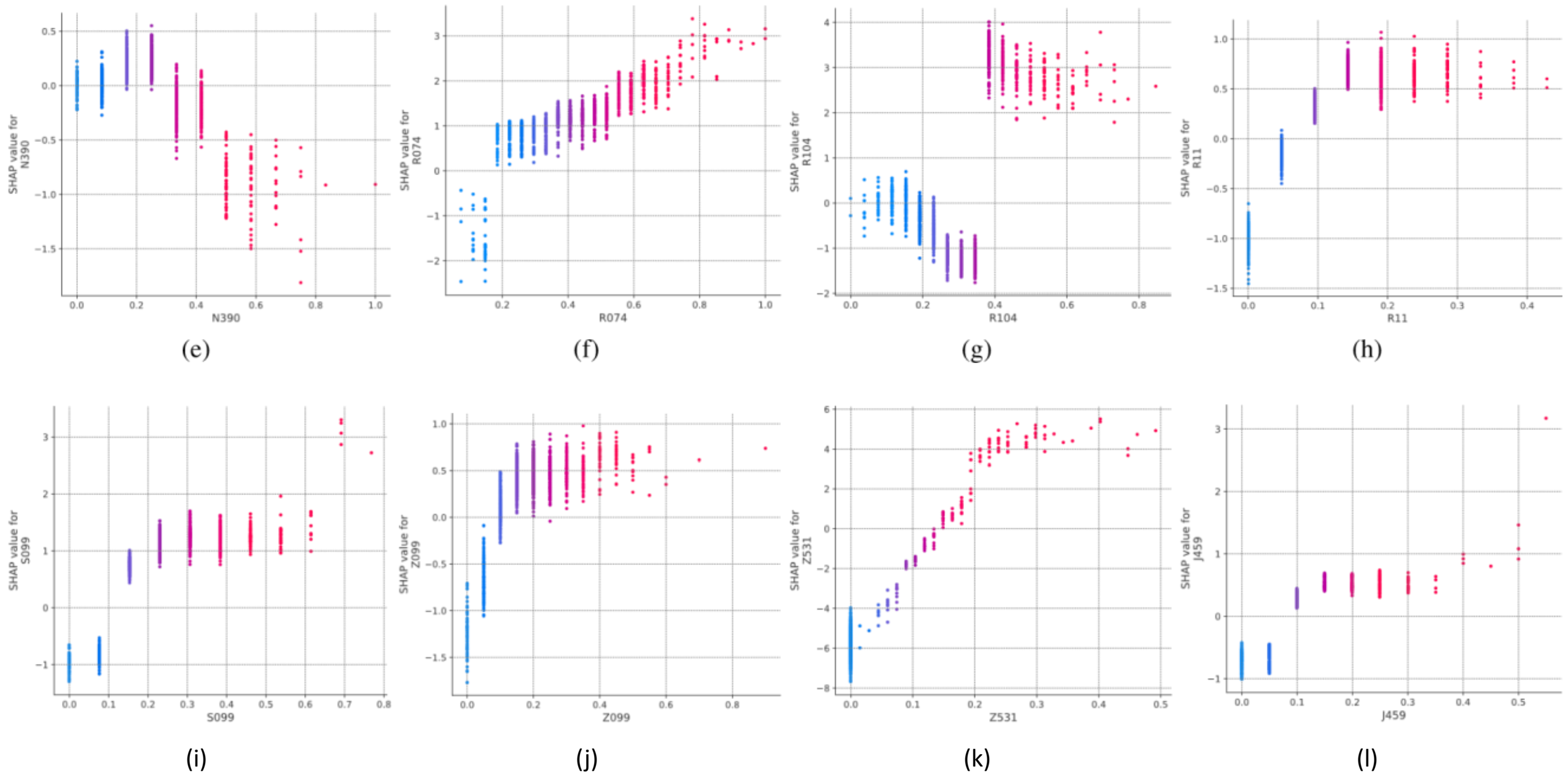

**Figure 19: Assessing the significance of the most prevalent ICD-10 diagnoses in forecasting the daily volume of ED visitors through the application of SHAP values.**

Finally, Figure A.4 showcases the significant contributions of the top five most frequently occurring ICD-10 codes documented daily. It is noted that elevated values of these features positively influence prediction outcomes, while lower values have a corresponding negative impact.

## 4. Conclusions

In this study, we address the limitations of existing AI-driven frameworks by introducing a novel Meta-learning Gradient Booster (Meta-ED) approach for accurately forecasting daily emergency department (ED) visits. Leveraging a comprehensive dataset of exogenous variables that include socio-demographic characteristics, healthcare service usage, chronic conditions, diagnoses, climate and temporal variables spanning 23 years from Canberra Hospital in the ACT, Australia, Meta-ED aims to enhance predictive accuracy and robustness for ED attendance patterns.

The Meta-ED model comprises four foundational learners—CatBoost, Random Forest, Extra Trees, and LightGBM—integrated with a reliable master learner, MLP. By combining the unique capabilities of various base models (sub-learners), Meta-ED strengthens the precision and reliability of its predictions. We evaluate the effectiveness of Meta-ED through a comprehensive comparative analysis involving 23 distinct models, encompassing diverse architectures such as sequential models, neural networks, tree-based algorithms, and ensemble methods.

The results indicate that Meta-ED significantly outperforms other models, achieving an accuracy of 85.7% (95% CI [85.4%, 86.0%]) across ten evaluation metrics. Compared with leading techniques, including XGBoost, Random Forest, AdaBoost, LightGBM, and Extra Trees, Meta-ED shows substantial accuracy improvements of 58.6%, 106.3%, 22.3%, 7.0%, and 15.7%, respectively. Additionally, incorporating weather-related features results in a 3.25% improvement in prediction accuracy, emphasising the model's ability to capture seasonal trends that affect patient volumes.

These promising results establish Meta-ED as a foundational model for precise prediction of daily ED visits, demonstrating its potential as a reliable and robust framework for predictive analytics in dynamic healthcare environments. Meta-ED offers a promising approach to improving prediction accuracy, particularly for sequential data characterised by heterogeneous features and limited datasets.

### Ethics approval

This study has received ethical approval from The ACT Health Human Research Ethics Committee (Approval No. 2024/ETH01037).

### Data availability

The Emergency Department (ED) Presentation dataset utilised in this study is proprietary to Canberra Health Services and the ACT Health Directorate. Data availability for our research is contingent upon accessing anonymised datasets, a process facilitated through coordination with Michael Phipps (michael.phipps@act.gov.au), the esteemed Senior Director of the Canberra Health Service. By engaging with him, researchers can access the requisite anonymised data sets for conducting comprehensive and insightful analyses.

### Declaration of conflicting interests

The authors declared no potential conflicts of interest with respect to the research, authorship, and/or publication of this article.

## Appendix A. Supplementary materials

### *Appendix A.1. Supplementary Tables*

**Table A.1: List of variables used in predicting the daily ED visitor number**

| Feature group | Feature name | Range |
|---|---|---|
| Calender data | Year<br>Month<br>Day<br>Weekday | 1999 to 2022<br>1 to 12 month of year<br>1 to 29 or 30 or 31, day of month<br>1 to 7 (Monday to Sunday) |
| Climate parameters | Daily Temperature | Daily Min Temp (degC) [-8.1, 26.7], μ = 6.94<br>Daily Max Temp (degC) [3.6, 43.65], μ = 20.77<br>Daily Mean Temp (degC) [0.5, 33.5], μ = 13.88 |
| | Daily Mean Humidity (%)<br>Daily Mean Precipitation (mm)<br>Speed Of Maximum Wind Gust (km/h)<br>Direction Of Maximum Wind Gust (Degrees) | [19.43, 99.75], μ = 68.27<br>[0.0, 84.0], μ = 1.76<br>[9.4, 105.5], μ = 37.76<br>[1, 360] |
| | Wind Speed (km/h) | [0.45, 42.8] |
| | Wind Direction | Wind Direction at 00:00 [0, 360]<br>Wind Direction at 06:00 [0, 360]<br>Wind Direction at 12:00 [0, 360]<br>Wind Direction at 21:00 [0, 360] |
| | Season | 1 to 4 |
| Gender | Female/ Male | 1 and 0 |
| Triage | Triage 1, Triage 2,..., Triage 5 | 1 to 5 |
| Age | Age 10 (%)<br>Age 10 20 (%)<br>Age 20 30 (%)<br>Age 30 40 (%)<br>Age 40 50 (%)<br>Age 50 60 (%)<br>Age 60 70 (%)<br>Age 70 (%) | [0.18, 35.34], μ = 18.11<br>[2.01, 28.90], μ = 12.90<br>[4.43, 33.02], μ = 15.83<br>[3.92, 23.30], μ = 12.84<br>[2.6, 20.86], μ = 10.57<br>[0.7, 20.71], μ = 9.17<br>[0.0, 18.90], μ = 7.37<br>[1.85, 26.06], μ = 13.20 |
| Five most frequent ICD-10 | ICD 1, ICD 2,..., ICD 5<br>ICD 1 N, ICD 2 N,..., ICD 5 N | [3, 67], μ = 14.11; [2, 25], μ = 8.38; [2, 18], μ = 6.37; [2, 13], μ = 5.01; [2, 11], μ = 4.27 |

**Table A.2: List of variables used in predicting the daily ED visitor number**

| Feature group | Feature name | Range |
|---|---|---|
| Disposition (%) | Dis Admit | [4.93, 51.08], μ = 31.15 |
| | Dis Home | [34.54, 90.35], μ = 59.07 |
| | Dis DNW | [0, 30.67], μ = 6.78 |
| | Dis TC | [0, 9.76], μ = 1.87 |
| | Dis LOR | [0, 4.72], μ = 0.42 |
| | Dis other | [0, 6.6], μ = 0.54 |
| Popular diagnosis (daily) | COPD | [0, 5], μ = 0.5 |
| | Asthma | [0, 20], μ = 2.19 |
| | Heart Failure | [0, 5], μ = 0.6 |
| | Hypertensive | [0, 5], μ = 0.3 |
| | Cardiovascular | [0, 22], μ = 6.65 |
| | Endocrine Nutritional Metabolic | [0, 9], μ = 1.5 |
| | Diabetes Mellitus | [0, 5], μ = 0.41 |
| | Mental _Behaviour | [0, 21], μ = 5.5 |
| | Genitourinary | [0, 26], μ = 6.9 |
| | Renal _Failure | [0, 4], μ = 0.38 |
| | Intentional Self Harm | [0, 6], μ = 0.18 |
| | Assault | [0, 6], μ = 0.17 |
| Most frequent ICD-10 (daily) | R074 | [0, 27], μ = 7.18 |
| | R10.4 | [0, 26], μ = 7.13 |
| | B34.9 | [0, 25], μ = 3.5 |
| | J45.9 | [0, 20], μ = 2.17 |
| | N39.0 | [0, 12], μ = 2.0 |
| | R11 | [0, 21], μ = 1.87 |
| | M54.5 | [0, 9], μ = 1.6 |
| | Z53.1 | [0, 67], μ = 10.3 |
| | Z09.9 | [0, 20], μ = 3.4 |
| | S09.9 | [0, 13], μ = 1.8 |

**Table A.3: Most common causes to visit ED based on whole visitors**

| Diagnosis | ICD10 code |
|---|---|
| Did not wait for treatment | (Z53.1) |
| Chest pain, unspecified | (R07.4) |
| Abdominal pain | (T79.4) |
| Suicidal ideation | (R45.81) |
| Pain in abdomen, other (includes colic) | (R10.4) |
| Viral infection, unspecified | (B34.9) |
| For review | (Z09.9) |
| Injury, unspecified or suspected of head | (S09.9) |
| Open wounds, lacerations - finger - uncomplicated | (S61.0) |
| Urinary tract infection | (N39.0) |
| Syncope or collapse (except heat syncope) | (R55) |
| Sprain and strain of ankle, part unspecified | (S93.40) |
| Nausea and vomiting (except in pregnancy) | (R11) |
| Asthma, Acute | (J45.9) |
| Low back pain | (M54.5) |
| Headache | (R51) |
| Fever | (R50.9) |
| Pneumonia, unspecified | (J18.9) |
| Multiple trauma | (T79.4) |
| Upper respiratory tract infection (URTI), acute | (J06.9) |
| Gastroenteritis and colitis of unspecified origin | (A09.9) |
| Tonsillitis | (J03.9) |
| Stroke, not specified as haemorrhage or infarction | (I64) |
| Migraine | (G43.9) |
| Sepsis (except with notifiable agent) | (A41.9) |

## *Appendix A.2. Supplementary figures*

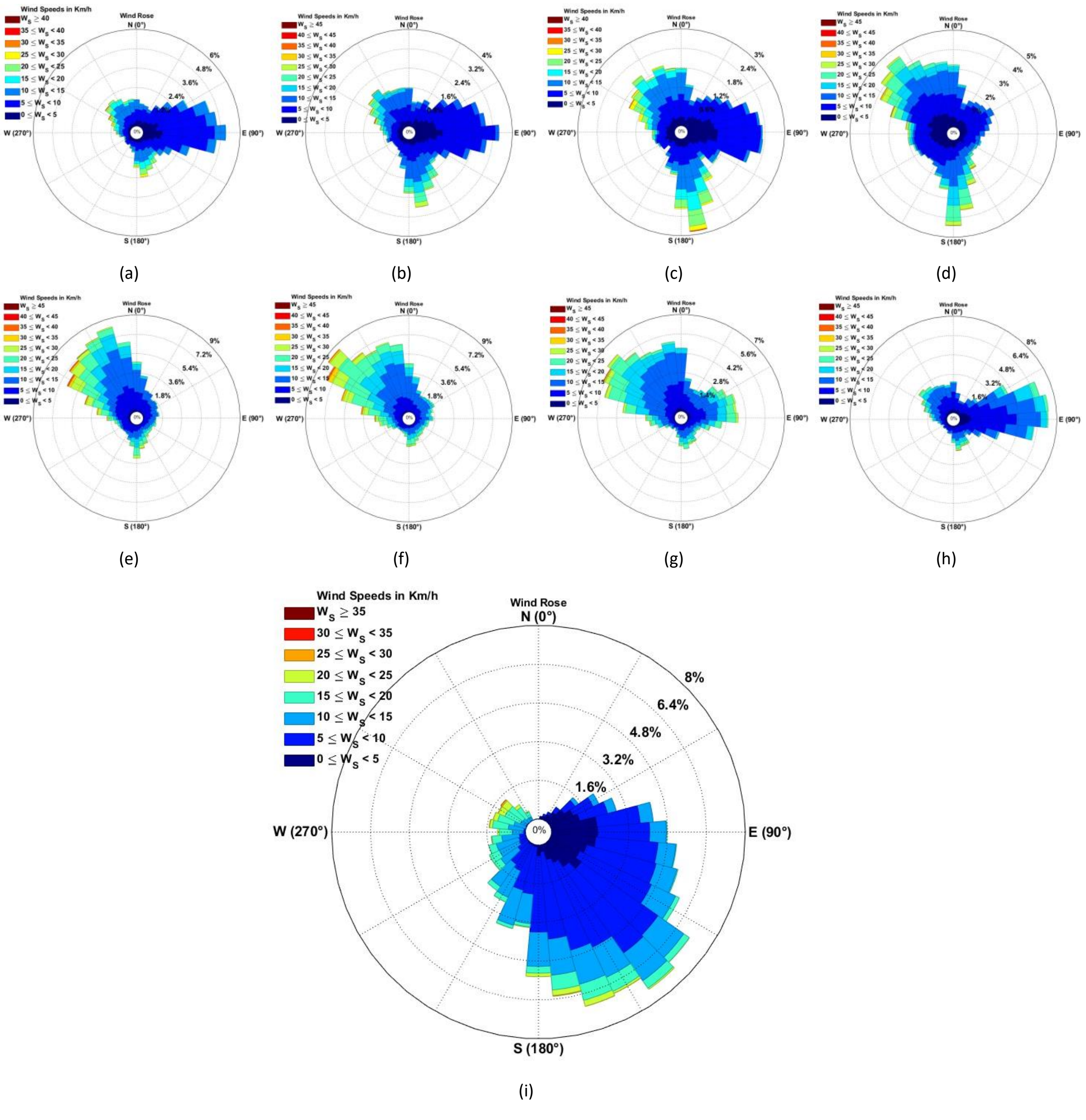

**Figure A.1: Wind roses show the temporal analysis of daily average wind speed (Km/h) and wind direction at Tuggeranong (Isabella Plains, Station ID: 070339) Canberra spanning 1996 to 2023, depicted in 3-hour intervals from midnight (a) to 9:00 PM (h).**

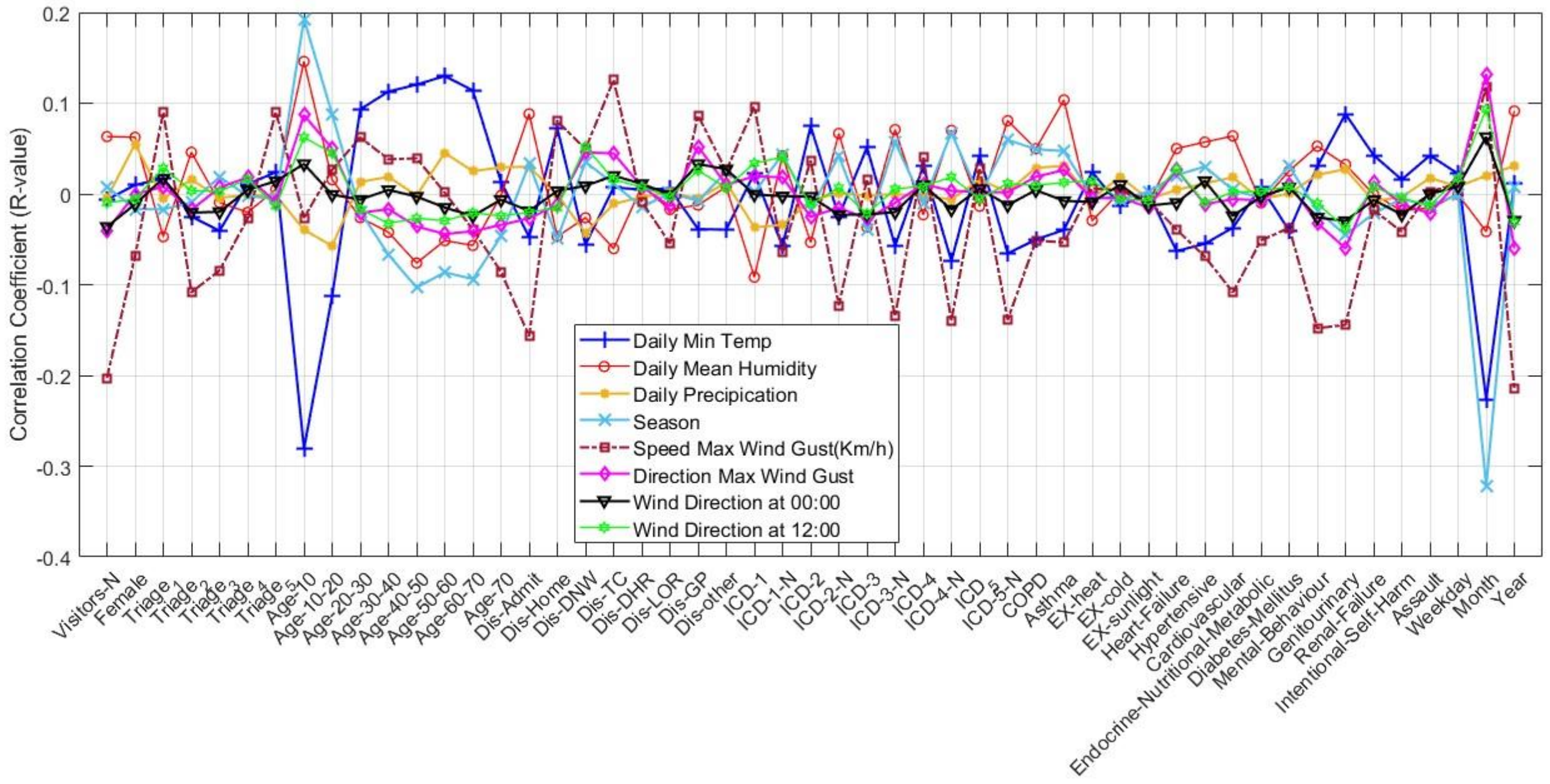

**Figure A.2: The correlations between climate variables and the ED visitors' number, patient demographic, medical information, and admission details.**

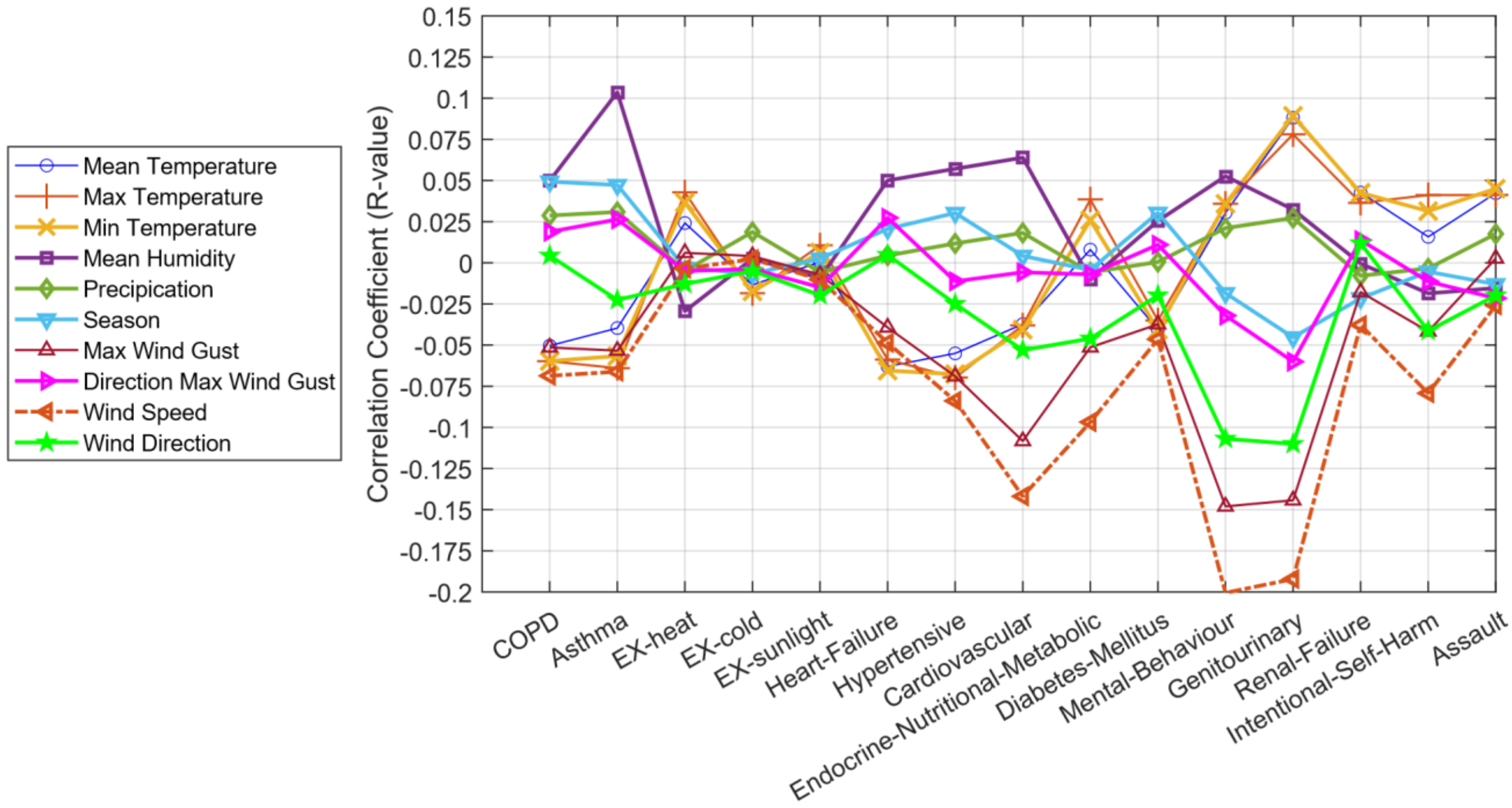

Figure A.3: The correlation between climate parameters and most frequent diagnosis based on ICD-10.

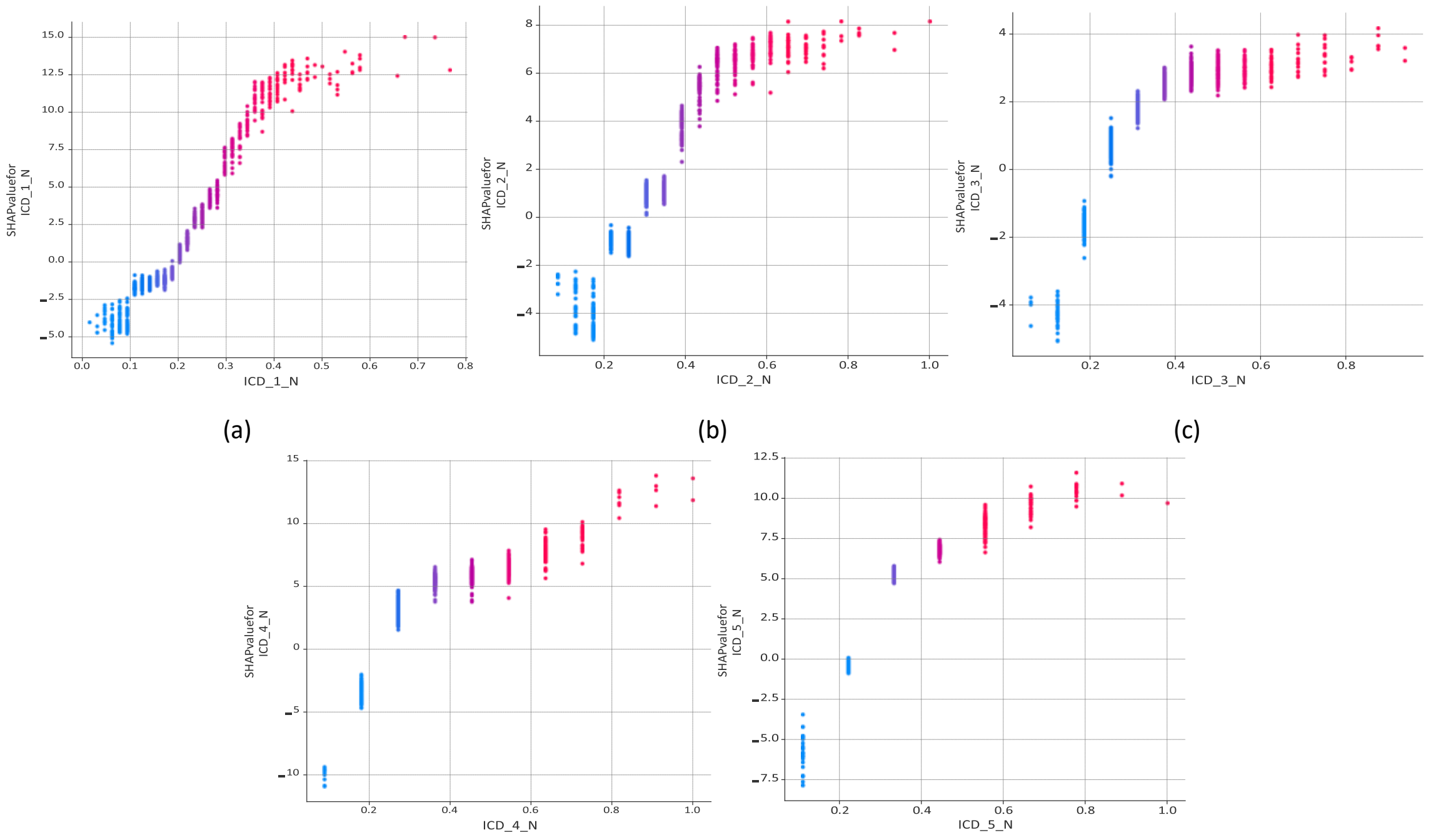

(a) (b) (c)

(d) (e)

**Figure A.4: Analysing the impact of the five most commonly encountered ICD-10 diagnoses on predicting the daily influx of ED visitors using SHAP values.**